\documentclass[acmsmall]{acmart}
\newif\ifextended
\extendedtrue

\AtBeginDocument{%
  }

\usepackage{cleveref}

\usepackage{amsmath}
\usepackage{amsthm}
\usepackage{enumitem}
\usepackage{tabularx}
\usepackage{booktabs}
\usepackage{subcaption}
\usepackage{multirow}
\usepackage{adjustbox}

\usepackage{xcolor}

\newif\ifshowrevisions
\showrevisionsfalse 

\newcommand{\rev}[1]{%
  \ifshowrevisions
    {\color{blue}#1}%
  \else
    {#1}%
  \fi
}

\usepackage[normalem]{ulem}

\usepackage{booktabs}
\usepackage{tabularx}
\usepackage{array}

\newcolumntype{Y}{>{\raggedright\arraybackslash}X}

\newcommand{\delrev}[1]{%
  \ifshowrevisions
    {\color{red}\sout{#1}}%
  \fi
}

\usepackage[noend]{algpseudocode}
\usepackage{algorithm}
\usepackage{siunitx}

\setcopyright{cc}
\setcctype{by}
\acmDOI{10.1145/3871753}
\acmYear{2027}
\acmJournal{PACMPL}
\acmVolume{11}
\acmNumber{POPL}
\acmArticle{51}
\acmMonth{1}
\acmSubmissionID{popl27main-p452-p}
\received{2026-07-09}
\received[accepted]{2026-10-05}



\begin{document}

\title{Solving Partially Observable Safety Games via Symbolic Memory Abstractions}

\author{Jia Pan}
\orcid{0009-0009-5010-870X}
\affiliation{%
  \institution{University of Texas at Austin}
  \city{Austin, TX}
  \country{USA}
}
\email{jia.pan@utexas.edu}

\author{Ziteng Wang}
\orcid{0009-0001-8487-8093}
\affiliation{%
  \institution{University of Texas at Austin}
  \city{Austin, TX}
  \country{USA}
}
\email{ziteng@utexas.edu}

\author{Venkat Arun}
\orcid{0000-0003-2192-1469}
\affiliation{%
  \institution{University of Texas at Austin}
  \city{Austin, TX}
  \country{USA}
}
\email{venkat@utexas.edu}

\author{Kenneth L. McMillan}
\orcid{0009-0000-9380-1939}
\affiliation{%
  \institution{University of Texas at Austin}
  \city{Austin, TX}
  \country{USA}
}
\email{kenmcm@cs.utexas.edu}

\author{Işıl Dillig}
\orcid{0000-0001-8006-1230}
\affiliation{%
  \institution{University of Texas at Austin}
  \city{Austin}
  \country{USA}
}
\email{isil@cs.utexas.edu}


\newcommand{\reduceSpacing}{
    \renewcommand{\baselinestretch}{0.95}
    
    \setlength{\parskip}{0.5em}

    \titlespacing{\section}{0pt}{*0.5}{*0.3}
    \titlespacing{\subsection}{0pt}{*0.4}{*0.2}
    
    \setlist{nosep} %
    
    \setlength{\footnotesep}{0.5em}
}

\newcommand{\tool}{\textsc{Verena}}
\newcommand{\todo}[1]{{\color{red}#1}}

\newcommand{\idiotbox}[2]{%
\vspace{-0.7em}
  \makebox[\linewidth][c]{%
    \fbox{%
      \begin{minipage}{0.93\linewidth}
        \textbf{Result for #1:} #2
      \end{minipage}
    }%
  }%
}

\newcommand{\bfpara}[1]{\vspace{5pt}\noindent\emph{\textbf{#1}}}

\newcommand{\LastObs}{\mathit{lastObs}}

\newcommand{\rulesize}{\fontsize{8.5pt}{10pt}\selectfont}

\newcommand{\States}{S}
\newcommand{\Actions}{A}
\newcommand{\Obs}{O}

\newcommand{\Init}{\mathcal{I}}         %
\newcommand{\Safe}{\phi}          %
\newcommand{\Model}{\mathcal{T}}        %

\newcommand{\Strat}{\mathcal{S}}           %
\newcommand{\Mem}{M}                       %

\newcommand{\Inv}{\mathsf{Inv}}   
\newcommand{\Reach}{\mathsf{Reach}} 
\newcommand{\ObsFunc}{\mathsf{Obs}}
\newcommand{\game}{\mathcal{G}}
\newcommand{\Vars}{\mathcal{V}}

\newcommand{\MemInit}{m_{\mathsf{init}}}
\newcommand{\MemUpdate}{\mathsf{upd}}
\newcommand{\Policy}{\pi}

\newcommand{\SafeReach}{\Psi}
\newcommand{\WinningMem}{\Phi}
\newcommand{\ComputeSafeReach}{\textsc{InferMemory}}
\newcommand{\SolveQuotient}{\textsc{SolveMemoryGame}}
\newcommand{\SolveMemory}{\textsc{SolveMemoryGame}}
\newcommand{\ConstructMemory}{\textsc{ConstructMemoryGame}}

\newcommand{\Enabled}{\mathsf{Enabled}}
\newcommand{\Update}{\mathsf{Update}}

\newcommand{\BlockedNext}{\mathsf{Bl}}

\newcommand{\Cand}{\mathit{Cand}}   %
\newcommand{\Next}{\mathit{Next}}   %
\newcommand{\Skel}{\mathit{Skel}}       %
\newcommand{\Guard}{\mathit{Guard}}     %
\newcommand{\Child}{\mathit{Child}}     %

\newcommand{\ExploreBeliefs}{\textsc{ExploreBeliefs}}
\newcommand{\Abstract}{\textsc{Abstract}}
\newcommand{\MineTemplates}{\textsc{MineTemplates}}

\newcommand{\BelHist}{B_{\mathit{hist}}}
\newcommand{\ValidHist}{H_{\mathit{valid}}}
\newcommand{\BadHist}{H_{\mathit{bad}}}

\newcommand{\isilsize}{\fontsize{8}{9}\selectfont}

\newcommand{\Templates}{\mathcal{T}}

\newcommand{\Covered}{\mathit{Covered}}
\newcommand{\FindUncoveredHist}{\mathsf{FindUncovered}}
\newcommand{\GeneralizeBelief}{\mathsf{GeneralizeBelief}}

\newcommand{\LocPolicy}{\mathit{LocPolicy}}
\newcommand{\Blocked}{\mathit{Blocked}}
\newcommand{\round}{r}

\newcommand{\Act}{\alpha}
\newcommand{\NextMem}{\beta}
\newcommand{\Updates}{U}
\newcommand{\ohist}{\omega}     
\newcommand{\Beliefs}{\mathit{Beliefs}}   %
\newcommand{\SafeSucc}{\textsc{SafeSucc}}

\newcommand{\PolicySeq}{\mathit{Policy}} %

\newcommand{\FeasibleHistories}{\textsc{FeasibleHistories}}
\newcommand{\SynthesizeLocalPolicy}{\textsc{LocalPolicy}}
\newcommand{\IsTotalOn}{\textsc{IsTotalOn}}
\newcommand{\PostBelief}{\textsc{PostBelief}}
\newcommand{\ExtractBadHistories}{\textsc{BadHistories}}

\newcommand{\Failure}{\textsc{Failure}}
\newcommand{\Unsat}{\textsc{Unsat}}

\newcommand{\FindUncovered}{\mathsf{FindUncovered}}

\newcommand{\QEexists}[1]{\textsc{QE}^\exists_{#1}}
\newcommand{\QEforall}[1]{\textsc{QE}^\forall_{#1}}

\newcommand{\commentsize}{\fontsize{8pt}{9pt}\selectfont}

\newcommand{\mycomment}[1]{\textcolor{blue}{\# \RaggedRight \texttt{\commentsize #1}}}
\newcommand{\constrain}{=\joinrel=}
\newcommand{\assert}{\texttt{\small assert}} 
\newcommand{\assume}{\texttt{\small assume}}

\newcommand{\irule}[2]%
   {\mkern-2mu\displaystyle\frac{#1}{\vphantom{,}#2}\mkern-2mu}
\newcommand{\irulelabel}[3]
{
\mkern-2mu
\begin{array}{ll}
\displaystyle\frac{#1}{\vphantom{,}#2} & \!\!\!\! #3
\end{array}
\mkern-2mu
}

\newcommand{\isil}[1]{{\color{red}{#1}}}

\newcommand{\tony}[1]{{\color{blue}{#1}}}

\begin{abstract}
Partially observable safety games formalize reactive synthesis problems in which a program must make safe choices despite having incomplete information about the underlying state. In such games, correctness  often depends not only on the current observable state, but also on past observations. This makes strategy synthesis  very challenging, particularly in symbolic settings with infinite state and action spaces.

This paper presents a new algorithm for synthesizing winning strategies for such games. A key idea is to algorithmically separate \emph{what to remember} from \emph{how to act}. To this end, our algorithm first performs generalization from bounded executions to infer a candidate \emph{symbolic memory abstraction} that a successful strategy should use. Such an abstraction consists of a set of variables, along with an interpretation of those variables in terms of the latent state. With this abstraction in place, the second phase of the algorithm then constructs a perfect information game over memory values, which can be solved via symbolic fixpoint computation to obtain both the action policy and a suitable memory-update function.

We implemented this approach in a prototype tool called \tool\ and evaluated it on {30} benchmarks drawn from UAV and descent control,  robotics, and congestion control. These benchmarks involve both continuous state variables arising from system dynamics and infinite action spaces arising from real-valued control choices. Although these benchmarks are beyond the scope of  existing  general-purpose methods, \tool\ solves {29} of the {30} instances, either by synthesizing a winning strategy or proving unrealizability, with a median solving time of {15.6} seconds.   Furthermore, on finite-action variants where comparison with prior work is possible, \tool\ solves more instances and scales better as the action space grows.

\end{abstract}


\maketitle
\section{Introduction}\label{sec:intro}

Reactive synthesis aims to automatically generate correct-by-construction strategies from logical specifications. In safety-critical systems, such as autonomous vehicles, robotics, and networked control systems, a central objective is  to ensure that the synthesized strategy maintains a safety property regardless of how the environment behaves. This goal naturally leads to a game-theoretic formulation, wherein the synthesized system and its environment form the two players of an infinite-duration game. In this game, the system chooses actions, the environment chooses any successor
allowed by a transition relation, and the system wins if all resulting
executions remain safe. In this view, the synthesis problem reduces to deciding whether the system has a winning strategy for the safety objective and, if so, constructing such a strategy. 

For finite-state systems with perfect information, this game-theoretic view is well understood. Such games admit \emph{memoryless} winning strategies, meaning that the synthesized system can choose its action based {only} on the \emph{current} game state, without remembering the history of how that state was reached. Thus, the main algorithmic task is to identify the states from which the system can force the game to remain safe, and decades of work have yielded effective algorithms that underlie successful reactive synthesis tools~\cite{PnueliRosner,EhlersGR1,ChatterjeeHenzinger07}.

However, this simple picture changes considerably under partial observability. When the synthesized system cannot observe the full game state, even safety objectives may require memory. That is, the system must choose actions based on what it can infer from its observation history, rather than based just on the current observation. A classical solution is to transform the game into one whose states are \emph{beliefs}, where each belief records the set of concrete states that remain possible after the observations seen so far~\cite{Reif84}. In this transformed game, the system has perfect information about its current belief, even though it does not know the true concrete state. However, this construction comes at a significant cost: even in finite-state systems, the number of beliefs can be exponentially larger than the original state space, and, in symbolic infinite-state systems,  directly constructing the belief game is generally infeasible (e.g., there may be infinitely many distinct beliefs, and their symbolic representations may grow unboundedly with the observation history). 
Prior work~\cite{DimitrovaFinkbeinerFSTTCS08} aims to address the infinite state setting by using predicate abstraction but assumes that system actions are drawn from a finite set.

Unfortunately, this restriction rules out many natural synthesis problems. In cyber-physical and networked systems, the state often contains continuous quantities such as position, velocity, altitude, queue length, or rate, while the synthesized strategy may need to choose real-valued actions such as steering angles, accelerations, or altitude corrections. At the same time, these systems rarely expose their full state to the controller; instead, decisions must be made from noisy or partial observations. Thus, the synthesis procedure must handle partial observability, infinite state spaces, and infinite action spaces simultaneously.

This paper presents a new algorithm that targets this  challenging setting. Our central idea is to algorithmically separate the synthesis of a strategy’s memory representation from the synthesis of the strategy itself. Rather than constructing an explicit finite abstraction of the belief game, our method learns a \emph{symbolic memory abstraction} that summarizes the safety-relevant uncertainty induced by observation histories. \rev{Importantly, such an abstraction can represent an infinite
memory domain using a finite symbolic description, without
requiring each memory value to represent an exact belief.} Once such an abstraction has been learned, the remaining synthesis problem can be viewed as an ordinary safety game over memory values, where actions are synthesized symbolically rather than selected from a finite action set. The formal basis of this reduction is an \emph{epistemic certificate}, which ensures that each memory value correctly summarizes possible latent states and that the strategy can continue to maintain safety.

\begin{figure}[t]
  \centering
  \vspace{-0.2in}
  \includegraphics[width=\linewidth]{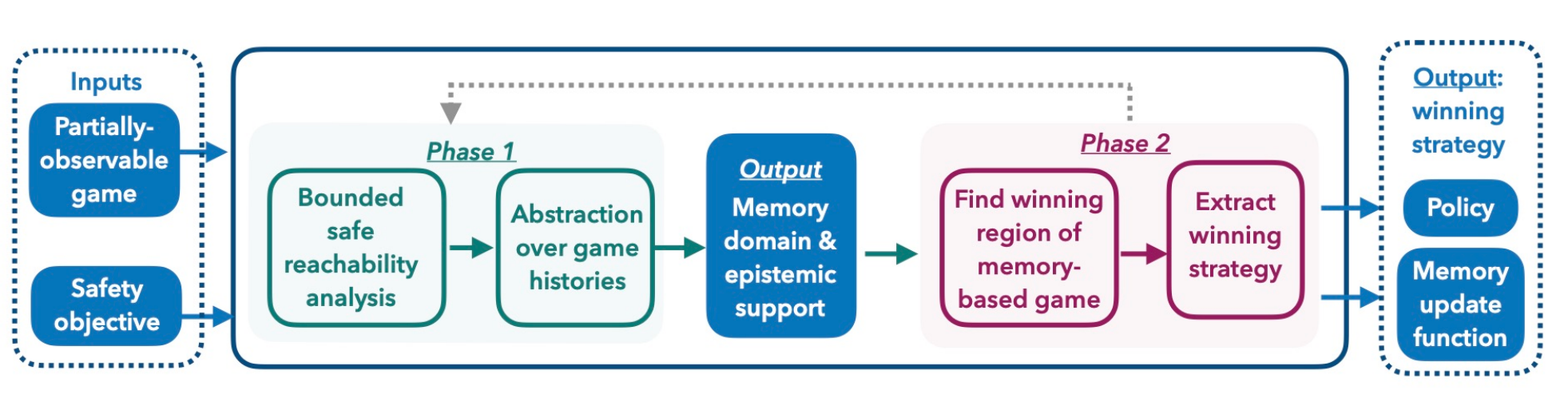}
  \vspace{-0.4in}
  \caption{High-level overview of our strategy synthesis algorithm. }
  \label{fig:overview}
\vspace{-0.27in}
\end{figure}

Figure~\ref{fig:overview} illustrates the resulting two-phase procedure. In the first phase, our algorithm analyzes safe finite executions to identify what information from the observation history is needed to make safe decisions. It then searches for recurring structure among the beliefs induced by these histories and generalizes them into a  {symbolic memory abstraction}, consisting of a set of \emph{memory variables} and an \emph{epistemic support relation} that gives meaning to memory values in terms of the latent states they represent.  The resulting abstraction gives the algorithm a compact symbolic representation of the controller’s uncertainty, together with the semantic information needed to reason about safety.

In the second phase, the algorithm uses this learned memory domain
to construct a perfect-information safety game over \emph{memory values}.
\rev{In each round of this game, the controller chooses an action,
the environment produces an observation, and the controller 
chooses a successor memory that soundly represents all possible
concrete successors. A memory value is \emph{winning} if the controller
can make these choices so as to maintain safety indefinitely.}
Solving the game therefore yields both the strategy's action choices
and its rule for updating its internal memory.
Crucially, the game is represented symbolically, so neither memory
values nor actions need to be enumerated. This allows our method
to reason directly about infinite memory domains as well as
infinite action spaces.

We have implemented our approach in a prototype tool, \tool, and evaluated it on a suite of partially observable safety games drawn from robot navigation, UAV altitude control,  network congestion control, and lander descent control. These benchmarks exercise the main features targeted by our approach and involve continuous state variables, partial observations, and real-valued action choices. On  {30} benchmarks with both infinite state and action spaces, \tool\ solves {29} instances, with a median solving time of {15.6} seconds. Since existing general-purpose techniques require finite action spaces, we additionally evaluate finite-action variants where comparison with prior work is possible.   In this setting, \tool\ solves more instances and scales better than prior work~\cite{DimitrovaFinkbeinerFSTTCS08} as the action space grows.

To summarize, this paper makes the following key contributions:
\begin{itemize}[leftmargin=*]
\item We introduce \emph{epistemic certificates}, a new proof principle for reasoning about memory-based strategies without explicitly constructing belief states. An epistemic certificate consists of an \emph{epistemic support relation}, which specifies the concrete states compatible
with each memory value, and a \emph{winning memory region}, which identifies
the memory values from which safety can be maintained indefinitely. \delrev{This decomposition enables strategy synthesis through
symbolic fixpoint reasoning over a perfect-information game.}
\rev{This framework enables strategy synthesis   by reducing the partially observable
game to a symbolic perfect-information game over memory values.}

\item Building on this representation, we develop an end-to-end synthesis algorithm
that learns a symbolic memory abstraction and its epistemic support relation
from bounded safe executions. It then constructs and solves the induced
perfect-information game to synthesize both a control policy and a memory-update
function.

\item We implement our approach in a new tool called \tool\ and evaluate it on benchmarks drawn from prior work. \tool\ handles partially observable games with both infinite state and action spaces and outperforms prior work on finite-action variants of the same benchmarks.
\end{itemize}

\section{Motivating Example}
\label{sec:overview}

\begin{figure}
  \centering

  \begin{subfigure}{0.32\linewidth}
    \centering
    \includegraphics[width=\linewidth]{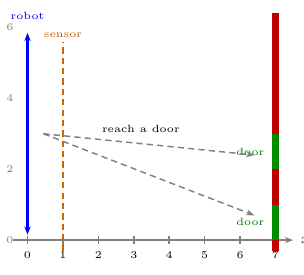}
    \vspace{-0.2in}
    \caption{Navigation game.}
    \label{fig:motivating-game}
  \end{subfigure}
  \hfill
  \begin{subfigure}{0.32\linewidth}
    \centering
    \includegraphics[width=\linewidth]{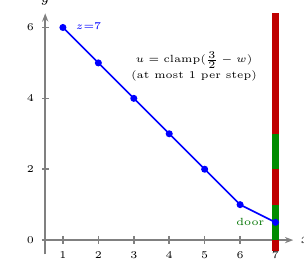}
    \vspace{-0.2in}
    \caption{Winning strategy.}
    \label{fig:motivating-strategy}
  \end{subfigure}
  \hfill
  \begin{subfigure}{0.32\linewidth}
    \centering
    \includegraphics[width=\linewidth]{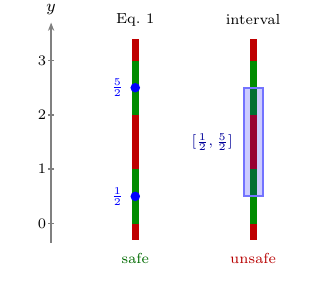}
    \vspace{-0.2in}
    \caption{Two different supports.}
    \label{fig:motivating-support}
  \end{subfigure}

\vspace{-0.13in}
  \caption{
  Motivating example. 
  (a) The robot must cross the wall through one of two doors while observing its
  altitude only once through a biased beacon.
  (b) The winning strategy remembers the beacon value $w$ and steers it toward
  a target interval.
  (c) The epistemic support explains what a memory value represents.
  }
  \label{fig:motivating}
\vspace{-0.30in}
\end{figure}

We illustrate the main ideas behind our approach using the simple navigation scenario
shown in Figure~\ref{fig:motivating-game}.  A robot (which we treat as a point for simplicity) moves from left to right
toward a vertical wall.  The wall has two narrow doors: a lower door at altitudes
in $(0,1)$, and an upper door at altitudes in $(2,3)$.  The
robot moves one unit to the right at every step, and it can choose a real-valued vertical
adjustment $u \in [-1,1]$ in each step.  It starts at horizontal position $x=0$
with an unknown altitude in $[0,6]$, and the wall lies at $x=7$.
The objective of this infinite-duration
safety game is to avoid colliding with the wall. Thus, when the robot reaches
$x=7$, it must cross through one of the two doors.

The difficulty is that the robot does not directly observe its altitude.  It
always observes its horizontal position $x$, but it receives altitude information
only once, from a localization beacon located at $x=1$.  The beacon reports a
value
$ 
    z = y + b,
$
where $y$ is the robot's true altitude and $b \in \{-1,+1\}$ is an unknown
sensor bias.  The bias is fixed throughout the execution, but the robot never
learns which value it has.  Hence, the same beacon reading can correspond to two
different physical situations.  For example, if the beacon reports $z=3/2$, then
the robot may be at altitude $1/2$ with bias $+1$, or at  $5/2$ with
bias $-1$.  Both possibilities remain indistinguishable to the robot.

While intentionally simple, this example exhibits several key features that are
of interest in this paper. First, the game is \emph{partially observable} because
the robot does not know its exact $y$-coordinate. Second, both the state and
action spaces are infinite: the altitude $y$ ranges over the reals, and the
controller chooses a real-valued adjustment $u$. Third, to achieve the game's
objective, the robot must use information from its observation history rather
than just its current observation. In particular, once the robot passes the
beacon, the current observation no longer contains any altitude information, but
the earlier beacon reading remains crucial for deciding how to steer.

\bfpara{A winning strategy.}
Despite the robot's uncertainty, this problem admits a simple winning strategy.
Although no memoryless strategy can win this game
\ifextended
(see Appendix~\ref{app:motivating}),
\else
(see the appendix of the extended version of the paper~\cite{extended_paper}),
\fi
the robot can ensure safety by retaining
the beacon reading observed at $x=1$ and updating it as it moves. Concretely,
when the robot reaches the beacon, it observes $z=y+b$ and initializes its internal memory, denoted $w$, to $z$. After each subsequent vertical adjustment $u$, it updates
its memory from $w$ to $w+u$. Since the bias $b$ is fixed, this maintains the relation
$w=y+b$ throughout the ``blind'' part of the execution, where the beacon reading is
no longer observable.

Now, equipped with $w$ as its memory, the robot can choose actions that are safe
for both possible values of the hidden bias. The strategy aims to make $w$ lie
in the interval $(1,2)$ by the time the robot reaches the wall. This is the
right target because, if $b=+1$, then the true altitude is $y=w-1$, while if
$b=-1$, then it is $y=w+1$. Thus, whenever $w\in(1,2)$, the two possible
altitudes lie in $(0,1)$ and $(2,3)$, which are exactly the two door intervals. The robot
therefore does not need to determine which bias is the true one; it only needs
to steer $w$ into this target interval.

The policy, shown in Figure~\ref{fig:motivating-strategy}, moves $w$
toward the midpoint $3/2$ as quickly as the action bound allows. Before reaching
the beacon, the robot chooses $u=0$, and, on subsequent steps, it chooses:
\begin{equation}\label{eq:strat}
u = \max(-1,\min(1,3/2-w))
\end{equation}

This strategy is winning because the robot has enough time to drive every possible beacon reading into the target interval: since $w \in [-1,7]$ after the beacon and the robot has six remaining vertical adjustments before reaching the wall, the policy above can ensure $w \in (1,2)$ by $x=7$.

\bfpara{Memory and epistemic support.}
The correctness argument above relies not only on the fact that the robot stores
a value $w$, but also on knowing what this value \emph{means}. After the beacon has
been observed, $w$ determines the following two possibilities for the latent
variables:
\begin{equation}\label{eq:epistemic}
  (b=+1 \wedge y=w-1)
  \;\vee\;
  (b=-1 \wedge y=w+1)
\end{equation}
This relation is ultimately what justifies the correctness of steering $w$ into the  $(1,2)$ range: for either possible value of $b$, $y$ will be either in the $(0, 1)$ or $(2, 3)$ range, so the robot will end up in one of the two doors. We refer
to relational invariants such as Eq.~\ref{eq:epistemic}, which give semantic
meaning to memory values, as \emph{epistemic support relations}.

In general, our algorithm need not recover the exact relation in Eq.~\ref{eq:epistemic}. It is sound to use an over-approximation, provided it contains all concrete states that may be compatible with a memory value. However, the over-approximation must still be precise enough to prove safety. For example, one could replace Eq.~\ref{eq:epistemic} with the coarser condition:
\begin{equation}\label{eq:interval-support}
 w-1 \leq y \leq w+1 
\end{equation}
This condition contains the two altitudes described by Eq.~\ref{eq:epistemic}, but it loses the information that only those two altitudes are possible. As shown in Figure~\ref{fig:motivating-support}, when $w=3/2$, Eq.~\ref{eq:interval-support} allows every altitude in $[1/2,5/2]$, including altitudes that hit the wall. By contrast, Eq.~\ref{eq:epistemic} allows only $1/2$ and $5/2$, both of which lie inside doors. Thus, epistemic support must preserve exactly those distinctions about the hidden state that matter for proving safety.

\bfpara{Algorithmic take-away.}
This example illustrates the two main tasks performed by our method. First, the
algorithm must discover what information should be retained in memory. In this
example, the relevant memory variable is $w$, whose meaning is given by the
support relation in Eq.~\ref{eq:epistemic}. Second, once this abstraction is
available, the algorithm must synthesize actions and memory updates that keep
the controller safe. Here, the action rule from Eq.~\ref{eq:strat} moves $w$ toward the target interval
$(1,2)$, and the memory update is simply $w := w+u$.
Our algorithm separates these tasks into two phases: in this example, the first
phase learns the memory variable $w$ together with its support relation in Eq.~\ref{eq:epistemic}, while
the second phase solves a symbolic perfect-information game over $w$ to
synthesize the policy in Eq.~\ref{eq:strat} and the update rule $w := w+u$.

\section{Preliminaries and Problem Statement}\label{sec:problem}

This work considers two-player safety games under partial observability, in which a \emph{safety player} interacts with an adversarial \emph{environment} that controls the system dynamics. Throughout this paper, we often use the word ``controller'' informally to refer to the safety player. The rest of this section  defines the game model, memory-based strategies, and the synthesis problem. 

\subsection{Partially Observable Safety Games }

To make the game model precise, we start with a symbolic representation of
states, actions, and observations. Let $\Vars_S$ be a finite tuple of state
variables and let $\Vars_A$ be a finite tuple of action variables, interpreted
over a background theory such as linear arithmetic over the reals or integers.
A state $s\in\States$ is a valuation of $\Vars_S$, and an action
$a\in\Actions$ is a valuation of $\Vars_A$. The sets $\States$ and $\Actions$
may be infinite. We model partial observability by distinguishing a sub-tuple
$\Vars_O \subseteq \Vars_S$ of observable state variables. An observation
$o\in\Obs$ is a valuation of $\Vars_O$, and the observation function
$ 
  \ObsFunc : \States \to \Obs
$
maps each concrete state to its projection onto the observable variables. We refer to state
variables not in $\Vars_O$ as \emph{latent}.

Throughout the paper, we write sets and relations symbolically as formulas over the underlying
background theory. In general, we write
$ 
  P(x,y,\ldots) \equiv \varphi(x,y,\ldots)
  \mid  x\in X,\; y\in Y,\ldots
$ 
to mean that $P$ denotes the set of tuples
$ 
  \{(x,y,\ldots) \mid x\in X,\; y\in Y,\ldots
     \wedge \varphi(x,y,\ldots)\}.
$
We omit the ranges when they are clear from context, and use primed variables
to denote next-state values.

\begin{definition}[Partially observable safety game]
A game instance
 $ \game $ is a triple $ (\Init,  \Model, \Safe)$
where $\Init(s)$ is a predicate characterizing the set of initial states $s \in \States$;  $\Safe(s)$ is a safety predicate over states; and  $\Model(s,a,s',o)$ is a transition relation describing the combined system and environment dynamics. That is, 
        given current state $s$ and  controller action $a$, the environment may
        choose any successor state $s' \in \States$ and observation $o \in \Obs$ such that $\Model(s,a,s',o)$ holds. We assume that $\Model$ is total and observation-consistent: for every
$s \in \States$ and $a \in \Actions$, some transition
$\Model(s,a,s',o)$ exists, and every such transition satisfies
$o=\ObsFunc(s')$.

\end{definition}

\begin{example}\label{ex:navigation-game}
Consider the navigation game from Section~\ref{sec:overview}. The concrete state
contains the robot's horizontal position $x\in\mathbb{R}$, true altitude
$y\in\mathbb{R}$, sensor reading $z \in \mathbb{R}$, and sensor bias
$b\in\{-1,+1\}$. Among these variables, $x$ and $z$ are observable, whereas
$y$ and $b$ are latent. The controller action is a real-valued vertical
adjustment $u\in[-1,1]$. The initial states are
\[
  \Init \equiv \bigl ( x=0 \wedge (0 \leq y \leq 6) \wedge
  (b = -1 \lor b = 1) \bigr )
\]
The safety condition requires that the robot has not collided with the wall:
\[
  \Safe \equiv \bigl( x = 7 \Rightarrow (0<y<1) \vee (2<y<3) \bigr)
\]
Finally, the  transition relation $\Model$ is given by:
\[
  x' = x+1
  \ \wedge\
  y' = y+u
  \ \wedge\
  b' = b
  \ \wedge\
  (x'=1 \Rightarrow z'=y'+b')
\]
\end{example}

\begin{definition}[Play]
A \emph{play} (or \emph{run}) of $\game$ is an infinite sequence of states $s_0s_1s_2\cdots$ together with actions $a_0,a_1,\ldots$ and observations $o_1,o_2,\ldots$ such that (1) $\Init(s_0)$ holds and, (2) for each $i\ge 0$,  we have $\Model(s_i,a_i,s_{i+1},o_{i+1})$.
A play is \emph{safe} if $\Safe(s_i)$ holds for all  $i \geq 0$.
\end{definition}

\subsection{Winning Strategies}

In partially observable games, the safety player cannot base its decisions on the
full state. Instead, it must choose actions using information obtained from the
observations seen so far. We model this information explicitly as an internal
\emph{memory}. A memory-based strategy updates its memory after each
observation and chooses actions as a function of the current memory value.

\begin{definition}[Memory-based strategy]
A \emph{memory-based strategy} is a tuple
$\Strat=\langle \Mem,\MemInit,\MemUpdate,\Policy\rangle$,
where $\Mem$ is a (possibly infinite) memory domain (e.g., $\mathbb{N} \times \mathbb{R}$), $\MemInit\in\Mem$ is the initial memory value,
$\MemUpdate:\Mem\times\Obs\to\Mem$ is a total memory update function, and
$\Policy:\Mem\to\Actions$ maps every memory value to an action.  Without loss of generality, we assume that each memory value includes the latest observation, so $\Policy$ may depend on current observations through memory.
\end{definition}

Once the controller commits to a strategy, only  plays that follow the
strategy's prescribed actions are allowed. We capture this using the following notion of an induced run:

\begin{definition}[Induced run]\label{def:plays}
Let 
$\Strat=\langle \Mem,\MemInit,\MemUpdate,\Policy\rangle$ be a strategy for game $\game=(\Init,\Model,\Safe)$. An \emph{induced run} of $\Strat$ in $\game$
is an infinite state sequence $(s_i)_{i\ge 0}$ such that  $ \Init(s_0)$  and 
there is a memory sequence $(m_i)_{i\ge 0}$ with $ m_0=\MemInit$ and:
\[ 
\forall i\ge 0.\ \bigl (   \ \Model (s_i,\Policy(m_i),s_{i+1},\ObsFunc(s_{i+1}))
\ \land
m_{i+1}=\MemUpdate(m_i,\ObsFunc(s_{i+1})) \ \bigr)
\]
\end{definition}

\begin{definition}[Winning strategy]\label{def:winning}
Let $\game=(\Init,\Model,\Safe)$ be a partially observable safety game. A strategy
$\Strat$ is \emph{winning} for $\game$ if every induced run $(s_i)_{i\ge 0}$ of
$\Strat$ in $\game$ is safe.
\end{definition}

\begin{example}\label{ex:navigation-strategy}
Consider again the navigation game from Example~\ref{ex:navigation-game}. As
discussed in Section~\ref{sec:overview}, the strategy remembers the beacon
reading, adjusted by the robot's own subsequent vertical moves. Since memory
values include the current observation, we take
$
  \Mem = \Obs \times \mathbb{R}$.
We write a memory value as $(o,w)$, where $o$ is the current observation
and $w$ is the remembered adjusted beacon value. For an observation $o$, we write
$o.x$ and $o.z$ for its position and beacon reading, respectively. The initial memory is
$
  \MemInit = (o_0,0),
$
where $o_0.x=0$ and $o_0.z=0$.
The memory update function copies the latest observation into memory and updates $w$ as follows:
\[
  \MemUpdate((o,w),o') =
  \begin{cases}
    (o',o'.z) & \text{if } o'.x=1 \\
    (o',w+\Policy(o,w)) & \text{otherwise }
  \end{cases}
\]
In other words, when the
robot  reaches the beacon, it records the observed beacon reading and then
 updates the remembered value according to the chosen vertical move.
The policy uses the current observed position $o.x$ and the remembered value $w$ to choose a vertical adjustment as follows:
\[
  \Policy(o,w) =
  \begin{cases}
    0 & \text{if } o.x < 1 \vee o.x \geq 7 \\
    \max(-1,\min(1,3/2-w)) & \text{if } 1 \leq o.x < 7
  \end{cases}
\]
Thus, before reaching the beacon and after crossing the wall, the controller chooses
$u=0$. Between the beacon and the wall, it moves $w$ toward the midpoint $3/2$
as much as the action bound allows.
\end{example}

\begin{definition}[Strategy synthesis]\label{def:problem}
Given a partially observable game $\game$, the \emph{strategy synthesis problem} is to either find a
winning strategy $\Strat$ for $\game$ or prove that no winning strategy exists.
\end{definition}

\section{Epistemic Certificates for Winning Strategies}
\label{sec:verification}

The previous section defines winning strategies semantically, by requiring all
induced runs to remain safe. However, for synthesis, we need a more local and symbolic
way to establish this property. The key idea is to give each memory value a
semantic interpretation as a set of concrete states it may represent, and then
reason inductively over memory values. 

This section formalizes this idea using \emph{epistemic certificates}. An
epistemic certificate has two components: an \emph{epistemic support relation},
which specifies the concrete states compatible with each memory value, and a
\emph{winning-memory predicate}, which identifies the memory values from which
safety can be maintained indefinitely. These components together provide a
local proof principle for memory-based strategies: a winning memory must represent
only safe concrete states, and it  must admit an action that, for every possible environment response, can be matched by an updated memory that (a) both soundly represents the resulting states and (b) remains winning.

\begin{definition}[Epistemic support relation]\label{def:epistemic-support}
An \emph{epistemic support relation} is a predicate
$\SafeReach(m,s)$ over memory values $m\in\Mem$ and concrete states
$s\in\States$. For a memory value $m$, the set
$ 
  \{\,s \mid \SafeReach(m,s)\}$
denotes the concrete states that are compatible with $m$.
\end{definition}

Intuitively, $\SafeReach(m,s)$ gives semantic meaning to memory values by specifying which concrete states may still be possible when the strategy's memory is $m$.

\begin{example}\label{ex:navigation-support}
Consider the navigation game from Section~\ref{sec:overview}. Recall that a
memory value has the form $m=(o,w)$, where $o$ is the current observation and
$w$ is the remembered adjusted beacon reading. As in
Example~\ref{ex:navigation-strategy}, we write $o.x$ and $o.z$ for the
components of the observation. Similarly, for a concrete state $s$, we write
$s.x$, $s.z$, $s.y$, and $s.b$ for its components.
For this example, a useful epistemic support relation  is the following:
\[
\begin{aligned}
  \SafeReach((o,w),s) \equiv{}&
    s.x=o.x
    \ \wedge\ (s.x =1 \Rightarrow s.z=o.z)
    \ \wedge\ \bigl(s.x=7 \Rightarrow (0<s.y<1) \vee (2<s.y<3)\bigr) \\
  &{}\wedge
    \bigl(
      o.x < 1
      \Rightarrow
      0 \le s.y \le 6
      \ \wedge\
      (s.b=1 \vee s.b=-1)
    \bigr) \\
  &{}\wedge
    \bigl(
      o.x \ge 1
      \Rightarrow
      (s.b=1 \wedge s.y=w-1)
      \ \vee\
      (s.b=-1 \wedge s.y=w+1)
    \bigr)
\end{aligned}
\]
Thus, before the beacon, $\SafeReach$ records the initial
uncertainty about the altitude and sensor bias. Once the current observation
indicates that the robot has reached or passed the beacon, the remembered value
$w$ precisely captures the two concrete altitudes compatible with the hidden
bias.
\end{example}

While the support relation gives each memory value a semantic interpretation,  it
does not by itself characterize the memories from which the controller can play
safely. We therefore pair it with a \emph{winning memory predicate}
$\WinningMem(m)$, identifying the memory values from which the strategy can
maintain safety. These two
predicates together form an \emph{epistemic certificate}:

\begin{definition}[Epistemic certificate]\label{def:epistemic-cert}
Let $\game=(\Init,\Model,\Safe)$ be a partially observable safety game, and let
$\Strat=\langle\Mem,\MemInit,\MemUpdate,\Policy\rangle$ be a memory-based strategy
for $\game$. An \emph{epistemic certificate} for $\Strat$ consists of a pair of
predicates $(\SafeReach,\WinningMem)$, where $\SafeReach(m,s)$ is an epistemic support relation and $\WinningMem(m)$ is a winning memory predicate satisfying the following conditions:
\begin{enumerate}[leftmargin=*]
\item {\bf (Init)} \quad 
$ 
\WinningMem(\MemInit)
\ \wedge
\forall s.\ (\Init(s)\Rightarrow \SafeReach(\MemInit,s))
$

\item {\bf (Safe)} \quad
$ 
\forall m,s.\
(\WinningMem(m)\wedge \SafeReach(m,s))
\Rightarrow \Safe(s)
$ 

\item {\bf (Step)} \quad
$ 
\forall m,o,s,s'.\quad
\WinningMem(m)\wedge \SafeReach(m,s)
\wedge \Model(s,\Policy(m),s',o)\
\Rightarrow
\WinningMem(\MemUpdate(m,o))
\wedge
\SafeReach(\MemUpdate(m,o),s')
 $
\end{enumerate}
\end{definition}

Condition (Init) says that the chosen initial memory must be winning (i.e., $\WinningMem(\MemInit)$) and must
cover all initial states (i.e., $\forall s.\ \Init(s)\Rightarrow \SafeReach(\MemInit,s)$). Next, condition (Safe) says that every state represented by
a winning memory is safe. Finally,  (Step) stipulates that, whenever the controller
acts according to $\Policy$ from a winning memory $m$, every successor state $s'$ chosen
by the environment is represented by the updated memory $\MemUpdate(m,o)$ (i.e., $\SafeReach(\MemUpdate(m,o),s')$) , and this
updated memory is again winning (i.e., $\WinningMem(\MemUpdate(m,o))$). \rev{Here, it is worth noting that the action $\pi(m)$ must work for every concrete
state represented by $m$, and the updated memory
$\mathrm{upd}(m,o)$ must represent every possible successor
consistent with observation $o$.}

\begin{theorem}[Certificate soundness]\label{thm:certificate-sound}
Let $\game=(\Init,\Model,\Safe)$ be a partially observable safety game, and let
$\Strat=\langle\Mem,\MemInit,\MemUpdate,\Policy\rangle$ be a memory-based strategy
for $\game$. If there exists an epistemic certificate
$(\SafeReach,\WinningMem)$ for $\Strat$, then $\Strat$ is winning for $\game$.%
\ifextended
\footnote{Proofs of all theorems are in the appendix.}
\else
\footnote{Proofs of all theorems are in the appendix of the
extended version of the paper~\cite{extended_paper}.}
\fi
\end{theorem}

\begin{example}
\label{ex:navigation-certificate}
Consider the strategy $\Strat$ from
Example~\ref{ex:navigation-strategy} and 
the support  $\SafeReach$ from Example~\ref{ex:navigation-support}. The epistemic certificate for $\Strat$ consists of $\SafeReach$ and   the following winning memory predicate $\WinningMem$:
\[
\begin{aligned}
\WinningMem(o,w) \equiv{}&
  (o.x=0)
  \ \vee\
  \bigl(
    1\leq o.x<7
    \ \wedge\
    o.x-6 < w
    \ \wedge\
    w < 9-o.x
  \bigr)
  \ \vee\
  (o.x\geq 7)
\end{aligned}
\]
Here, the middle disjunct characterizes memories from which the controller can still
force $w$ into the target interval $(1,2)$ before reaching the wall. If the
current observed position is $o.x$, then there are $d=7-o.x$ vertical
adjustments left. Since each adjustment can change $w$ by at most one, the
controller can increase $w$ by at most $d$ and decrease it by at most $d$ before
the wall is reached. Thus $w$ can still be driven into $(1,2)$ exactly when
$
  w+d>1$  and  $w-d<2$
or equivalently,
$
  1-(7-o.x) < w < 2+(7-o.x)
$, which simplifies to
$
  o.x-6 < w < 9-o.x$.
\end{example}

\section{Strategy Synthesis via Epistemic Certificates}
\label{sec:algorithm-overview}

In this section, we present our algorithm for finding a winning strategy for a partially observable game, along with an epistemic certificate validating the correctness of the synthesized strategy.

\subsection{Top-Level Algorithm}
Algorithm~\ref{alg:solve-po-game} summarizes our top-level algorithm.  As discussed in Section~\ref{sec:intro}, the key idea is to separate the problem of discovering what to
remember from that of deciding how to act. To this end, Algorithm~\ref{alg:solve-po-game} first tries to infer a
candidate memory abstraction consisting of memory variables and their corresponding  epistemic support. It then asks whether this abstraction contains enough information to
support a winning strategy.
To find the appropriate support relation, the algorithm
analyzes safe finite prefixes of the game, using a parameter 
 $h$ to bound the length of prefixes considered in each iteration.
Smaller horizons are easier to explore but may yield memory abstractions that
are too coarse. In contrast, larger horizons are more expensive to explore but provide more information. Therefore, the algorithm explores horizons in increasing
order, starting from a small initial value and increasing $h$ whenever the
current abstraction is insufficient. As shown in Algorithm~\ref{alg:solve-po-game}, each iteration of the {\sc FindWinning} involves
 three key sub-procedures: $\ComputeSafeReach$, $\ConstructMemory$, and $\SolveQuotient$.

\begin{algorithm}[t]
\caption{\textsc{FindWinning}$(\Init,\Model,\Safe)$}
\label{alg:solve-po-game}
\begin{algorithmic}[1]
\State $h \gets \mathit{InitHorizon}()$
\While{true}
  \State $(\Mem,\SafeReach) \gets \ComputeSafeReach(\Init,\Model,\Safe,h)$
  \If{$\Mem = \bot$}
     \Return \textsc{Fail}
  \EndIf
  \State $\game_{\SafeReach} \gets
    \ConstructMemory(\Mem,\SafeReach,\Init,\Model,\Safe)$
  \State $r \gets \SolveMemory(\game_{\SafeReach})$
  \If{$r = \mathit{Success}(\MemInit,\Policy,\MemUpdate)$}
     \Return $\langle \Mem,\MemInit,\MemUpdate,\Policy \rangle$
  \ElsIf{$r = \mathit{Failure}$}
     $h \gets \mathit{IncreaseHorizon}(h)$
  \EndIf
\EndWhile
\end{algorithmic}
\end{algorithm}

The first subroutine, \textsc{InferMemory} (line 3), is responsible for
inferring what the controller needs to remember. It does so by analyzing safe
finite game prefixes of length at most $h$ and summarizing the uncertainty
induced by their observation histories. The intuition is that many histories may
induce uncertainty with recurring symbolic structure, so the controller should
store a compact symbolic summary rather than the histories themselves.
Section~\ref{sec:forward} describes how the algorithm turns this
intuition into a symbolic memory domain and its corresponding support relation.

The second subroutine, $\ConstructMemory$ (line~5), uses the candidate epistemic
support relation $\SafeReach$ to construct a perfect-information game over
memory values, denoted $\game_{\SafeReach}$. Since $\SafeReach$ interprets each
memory value as a set of possible concrete states, the new game can account for
the original game dynamics \emph{without} making latent states part of the game state. In
this so-called \emph{memory game}, a round consists of three steps: the controller chooses an action from
the current memory, the environment chooses a compatible observation, and the
controller chooses a successor memory. Thus, $\game_{\SafeReach}$ is a perfect-information game whose states are memory values. 

Finally, $\SolveMemory$ solves the memory game $\game_{\SafeReach}$ (line~6).
Because this game is perfect-information, we can use standard safety-game
reasoning to compute its winning region. Intuitively, $\SolveMemory$ identifies
memory values from which the controller can choose an action such that every observation chosen by the environment can be followed by an update to another winning memory.
If this search succeeds, the algorithm chooses an initial memory satisfying the
memory game's initial condition and lying in the winning region, and extracts
witnesses for the policy $\Policy$ and memory update function $\MemUpdate$.

On the other hand,  if $\SolveMemory$ fails, this does not mean that the original game has no
winning strategy. Since the memory abstraction was inferred from bounded
prefixes, it may simply be too coarse to support an inductive certificate. The
algorithm therefore increases the horizon $h$ and repeats (line~8). A
larger horizon exposes longer observation histories and may either result in a 
more precise memory abstraction or reveal that the game is unrealizable.

The remainder of this section explains the three key subroutines of   Algorithm~\ref{alg:solve-po-game} in more detail.

\subsection{Induced Perfect-Information Game on Memories}
\label{sec:quotient-game}

In this subsection, we describe the construction of the perfect-information memory game for a given memory domain and its corresponding support.   Although this phase runs after
support construction, we present it first because it clarifies how the epistemic support is used.

The key idea is to make memory values the states of the new game. Because the
epistemic support gives each memory value its meaning in terms of concrete
states, the game can account for the original dynamics without making those
states an explicit part of the game. Thus, each round proceeds entirely at the
level of memories and observations: the controller chooses an action from the
current memory, the environment produces an observation for that action, and the controller chooses a
successor memory. \rev{Crucially, the successor memory is not uniquely determined
by the current memory, action, and observation. Instead, the
controller can choose any successor memory that soundly represents
all possible concrete successors, allowing the memory-update
function to be synthesized along with the control policy.} 
\delrev{Formally, we define the \emph{memory game } as follows:}

\begin{definition}[Induced memory game]
\label{def:memory-game}
Given a partially observable game $\game=(\Init,\Model,\Safe)$, a memory
domain $\Mem$, and an epistemic support relation $\SafeReach(m,s)$ such that
$\SafeReach(m,s)\Rightarrow \Safe(s)$, the \emph{memory game} induced by
$(\Mem,\SafeReach)$ is the round-based game
$ 
  \game_{\SafeReach}
  =
  (\Init_{\SafeReach}, \Model_{\SafeReach}, \Safe_{\SafeReach},
   \Enabled_{\SafeReach})
$ 
where: 
\begin{itemize} 
  \item $\Init_{\SafeReach}(m) \equiv
  \forall s.\ \Init(s)\Rightarrow \SafeReach(m,s)$;
  \item $\Enabled_{\SafeReach}(m,a,o) \equiv
  \exists s,s'.\ \SafeReach(m,s)\wedge \Model(s,a,s',o)$; 
  
  \item $\Model_{\SafeReach}(m,a,o,m') \equiv
  \forall s,s'.\
  \SafeReach(m,s)\wedge \Model(s,a,s',o)
  \Rightarrow \SafeReach(m',s')$; 

  \item $\Safe_{\SafeReach}(m) \equiv
  \exists s.\ \SafeReach(m,s)$.

\end{itemize}
A round from memory $m$ proceeds as follows: the controller chooses an action
$a$, the environment chooses an observation $o$ satisfying
$\Enabled_{\SafeReach}(m,a,o)$, and the controller then chooses a successor
memory $m'$ satisfying $\Model_{\SafeReach}(m,a,o,m')$.
\end{definition}

Here, the initial condition $\Init_{\SafeReach}$ requires the initial memory to
cover all possible initial concrete states.
\delrev{The transition relation
$\Model_{\SafeReach}$ identifies which successor memories are allowed after a
given action and observation. In particular, a successor memory $m'$ is allowed
only if it represents every concrete successor $s'$ reachable via a transition
that produces observation $o$ from some state $s$ represented by $m$.}
\rev{The relation $\Enabled_{\SafeReach}$ identifies which observations the
environment can produce after action $a$ from memory $m$. In contrast,
$\Model_{\SafeReach}$ determines which successor memories the controller
can choose after receiving an observation $o$. In particular, the chosen
memory $m'$ must represent every concrete successor consistent with
the current memory, action, and observation.}
\rev{Finally}, the safety condition $\Safe_{\SafeReach}$ requires the current memory to
represent at least one concrete state. Because $\SafeReach$ only relates
memories to safe states, i.e., $\SafeReach(m,s)\Rightarrow\Safe(s)$, this means
that the set of concrete states represented by $m$ is nonempty and contained in
$\Safe$.
\delrev{Finally, $\Enabled_{\SafeReach}$
identifies which observations the environment is allowed to produce after
action $a$ from memory~$m$.}
\rev{Furthermore, any action that can lead to an unsafe
concrete state is ruled out because no successor memory
can represent that state while satisfying the epistemic
support relation.}

Importantly, the resulting turn-based game is perfect-information because both
players have access to the full state of the induced game.
\delrev{The controller chooses an action based on the current memory value
$m$, which is visible. Similarly, when choosing a successor memory, the
controller has access to the current memory $m$, the action $a$ chosen
earlier in the round, and the observation $o$ produced by the environment.}
\rev{In particular, the states of this game are memory values, which are
known to the controller. The latent states of the original game appear
only as quantified variables in $\Enabled_{\SafeReach}$ and
$\Model_{\SafeReach}$, rather than as part of the game state.
Thus, the controller has all the information needed to choose both
its action and successor memory.}

\begin{example}
\label{ex:navigation-induced-game}
Consider the navigation game and support relation from
Examples~\ref{ex:navigation-support} and~\ref{ex:navigation-certificate}. The
memory game has states $(o,w)$ and actions $u\in[-1,1]$; in each round, the
environment produces a successor observation $o'$. The latent variables $y$ and
$b$ no longer appear as game states, as their possible values are accounted for
by $\SafeReach$.
The initial condition is
$\Init_{\SafeReach}(o,w) \equiv o.x=0$, and the memory-safety predicate $\Safe_{\SafeReach}(o, w)$ is given by:
\[
  o.x = 7 \Rightarrow
  ( (1<w<2) \vee (3<w<4) \vee (-1<w<0)  )
\]

The enabled observations are those that can arise from some concrete state. After eliminating the latent variables,
we obtain the following formula for $\Enabled_{\SafeReach}((o,w),u,o')$:
\[
\begin{aligned}
  \Safe_{\SafeReach}(o,w)
  \ \wedge\
  -1 \le u \le 1
  \ \wedge\
  o'.x=o.x+1  \wedge{} \bigl(o'.x=1 \Rightarrow u-1 \le o'.z \le u+7\bigr)
\end{aligned}
\]

\delrev{Finally, the transition relation
$\Model_{\SafeReach}((o,w),u,o',(o'',w'))$ is given by:}
\rev{Finally, for enabled observations, the induced transition
relation $\Model_{\SafeReach}$ simplifies to:}

\rev{
\[
\begin{aligned}
  o''=o'
  \ \wedge\
  \begin{cases}
    w'=o'.z & \text{if } o.x=0 \\
    w'=w+u & \text{if } 1\le o.x\le 5 \lor o.x \geq 7 \\
    w'=w+u \wedge 1<w'<2 & \text{if } o.x=6
  \end{cases}
\end{aligned}
\]
}
\end{example}

Recall that, in the induced memory game, the controller makes two kinds of choices: it first
chooses the system action, and, after the environment produces an observation,
it chooses the next memory. Thus, a strategy for the memory game has two
components:

\begin{definition}[Strategy of memory game]
\label{def:memory-strategy}
A strategy for $\game_{\SafeReach}$ is a pair $(\Policy,\MemUpdate)$ where
$\Policy: \Mem \to \Actions$ and
$\MemUpdate: \Mem \times \Obs \to \Mem$. Given a strategy $(\Policy,\MemUpdate)$ and a memory $m_0$, a finite prefix
generated by $(\Policy,\MemUpdate)$ is a sequence
$
  m_0,o_1,m_1,\ldots,o_k,m_k
$
such that, for every $0 \leq i < k$,
$\Enabled_{\SafeReach}(m_i,\Policy(m_i),o_{i+1})$ and
$m_{i+1}=\MemUpdate(m_i,o_{i+1})$. 
\end{definition}

The policy $\Policy$ resolves the controller's first choice,
namely which action to take from the current memory. The update function
$\MemUpdate$ resolves the controller's second choice, namely which successor
memory to use. Next, we define what it means for a strategy to be winning for our memory game:

\begin{definition}[Winning strategy of memory game]
\label{def:memory-winning-strategy}
A strategy $(\Policy,\MemUpdate)$ is \emph{winning from} a memory
$m_0 \in \Mem$ if, for every finite prefix
$
  m_0,o_1,m_1,\ldots,o_k,m_k
$
generated by $(\Policy,\MemUpdate)$, the following hold:
\begin{itemize}[leftmargin=*]
  \item \textbf{Safety.}
  $\Safe_{\SafeReach}(m_k)$ holds.
  \item \textbf{Update validity.}
  For every observation $o$ such that
  $\Enabled_{\SafeReach}(m_k,\Policy(m_k),o)$ holds, we have
  \[
    \Model_{\SafeReach}(m_k,\Policy(m_k),o,\MemUpdate(m_k,o)).
  \]
\end{itemize}
A memory $m_0$ is \emph{winning} if there exists a strategy that is winning
from $m_0$.
\end{definition}

Here, the first condition states that winning strategies always ensure safety. The second condition, called \emph{update validity}, requires the strategy's chosen memory
update to be a valid transition of the induced memory game. That is, whenever the
environment produces an enabled observation, the update function must choose a
successor memory allowed by $\Model_{\SafeReach}$. \rev{Without this requirement, the controller could choose a successor
memory that appears safe but fails to represent some of the concrete
states that may actually result from its action. Unlike in the
original game, where transition validity is built into the
definition of an induced run, we must therefore explicitly
require the controller's memory updates to be valid.}

The next theorem relates the induced memory game to the original partially
observable game:

\begin{theorem}[Correctness of the induced memory game]
\label{thm:quotient-correctness}
Let $\game=(\Init,\Model,\Safe)$ be a partially observable safety game, and let
$\SafeReach(m,s)$ be an epistemic support relation for memory domain $M$ satisfying $\SafeReach(m,s)\Rightarrow\Safe(s)$.  Finally let $\game_\SafeReach =  (\Init_{\SafeReach}, \Model_{\SafeReach}, \Safe_{\SafeReach},
   \Enabled_{\SafeReach}),$ be the memory game induced  by $\SafeReach$ on $\game$. We have the following two properties:
\begin{enumerate}[leftmargin=*]
  \item \textbf{Soundness.}
  Suppose $\MemInit$ satisfies $\Init_{\SafeReach}(\MemInit)$, and suppose
  $(\Policy,\MemUpdate)$ is winning from $\MemInit$ in
  $\game_{\SafeReach}$. Then
 $ 
    \Strat=\langle \Mem,\MemInit,\MemUpdate,\Policy\rangle
  $
  is a winning strategy for $\game$.

  \item \textbf{Relative completeness.}
  Suppose
  $
    \Strat=\langle \Mem,\MemInit,\MemUpdate,\Policy\rangle
  $
  is a strategy for $\game$ and there exists a memory predicate $\Phi(m)$ such
  that $(\SafeReach,\Phi)$ is an epistemic certificate for $\Strat$ and
  $\Phi(m)\Rightarrow\Safe_{\SafeReach}(m)$ for all $m$. Then
  $(\Policy,\MemUpdate)$ is winning from $\MemInit$ in
  $\game_{\SafeReach}$.
\end{enumerate}
\end{theorem}

The soundness part of this theorem states that a solution to the induced memory
game can be used to obtain a winning strategy for the original partially
observable game. Thus, for any epistemic support relation satisfying
$\SafeReach(m,s)\Rightarrow\Safe(s)$, we can first construct the induced memory
game and attempt to solve it. If a solution exists, then the resulting memory
policy and update function are guaranteed to be safe for the original game. 
Conversely, the relative completeness part of the theorem shows that the
reduction to a perfect-information memory game does not result in loss of
completeness relative to $\SafeReach$. That is, assuming a strategy $\Strat$ for the
original game admits an epistemic certificate using this particular support
relation $\SafeReach$, the corresponding strategy for $\Strat$ is also winning in the induced memory game. 

\subsection{Solving the Memory Game}

In the previous subsection, we showed that the reduction to the induced memory
game is correct. It now remains to compute a winning strategy for it. Fortunately, because
$\game_{\SafeReach}$ is a perfect-information safety game, we can use the standard
greatest-fixpoint characterization of its winning region. For a candidate memory region $\Phi$, we define the following \emph{controllable predecessor} operator, as standard in the analysis of safety games~\cite{deAlfaroHM01}:
\[
\mathsf{Pre}_{\SafeReach}(\Phi)(m)
\equiv
  \Safe_{\SafeReach}(m)
  \wedge
  \exists a.\ \forall o.\
    \Enabled_{\SafeReach}(m,a,o)
    \Rightarrow
    \exists m'.\
      \Phi(m') \wedge \Model_{\SafeReach}(m,a,o,m').
\]

\delrev{This operator defines safe memories from which the controller can force the
next memory into $\Phi$. The conjunct $\Safe_{\SafeReach}(m)$ requires the current
memory itself to be safe, and the remaining formula asks for an action $a$ such that,
for every observation $o$ the environment can produce, some successor memory $m'$
is both in $\Phi$ and a valid $\Model_{\SafeReach}$-transition.}
\rev{This operator characterizes the memories from which the controller can
force the next memory into $\Phi$. The conjunct $\Safe_{\SafeReach}(m)$
requires the current memory to be safe, while the remaining
existential-universal-existential quantification reflects the order of
choices in the induced game. Specifically, the controller chooses an
action $a$, the environment produces any enabled observation $o$, and
the controller then chooses a successor memory $m'$ that belongs to
$\Phi$ and satisfies $\Model_{\SafeReach}(m,a,o,m')$.}
Using this operator, we can now define the \emph{winning region} of the induced memory game as the greatest
fixpoint of $\mathsf{Pre}_{\SafeReach}$:
\[
  \WinningMem \triangleq \nu X.\ \mathsf{Pre}_{\SafeReach}(X).
\]
In other words, $\WinningMem$ is the largest set of memories from which the controller can
keep the induced game safe indefinitely.

\begin{algorithm}[t]
  \caption{\textsc{SolveMemoryGame}$(\game_{\SafeReach})$}
  \label{alg:solve-mem}
  \begin{algorithmic}[1]
    \State $\Cand(m) \gets \Safe_{\SafeReach}(m)$
    \While{true}
      \State $\Next(m) \gets
        \Safe_{\SafeReach}(m) \wedge
        \exists a.\ \forall o.\
          \Enabled_{\SafeReach}(m,a,o)
          \Rightarrow
          \exists m'.\
            \Cand(m') \wedge \Model_{\SafeReach}(m,a,o,m')$
      \If{$\forall m.\ \Cand(m) \Leftrightarrow \Next(m)$}
        \State $\WinningMem(m) \gets \Cand(m)$
        \If{$\neg \exists m.\ \WinningMem(m) \wedge \Init_{\SafeReach}(m)$}
     \Return $\mathit{Failure}$
        \EndIf
        \State $\MemInit \gets
          \textsc{GetModel}\bigl(\Init_{\SafeReach}(m) \wedge \WinningMem(m)\bigr)$
        \State $(\Policy,\MemUpdate) \gets
          \textsc{ExtractStrategy}(\game_{\SafeReach},\WinningMem)$
        \State \Return $\mathit{Success}(\MemInit,\Policy,\MemUpdate)$
      \EndIf
      \State $\Cand(m) \gets \Next(m)$
    \EndWhile
  \end{algorithmic}
\end{algorithm}

Algorithm~\ref{alg:solve-mem} computes this fixpoint and extracts a strategy
from the resulting winning region. The main loop iterates
$\mathsf{Pre}_{\SafeReach}$, starting from $\Safe_{\SafeReach}$, until it reaches
a fixpoint $\WinningMem$. If no initial memory lies in this region
(line~6), then the current support relation $\SafeReach$ admits no winning
initial memory, and the algorithm returns $\mathit{Failure}$. 
Otherwise, the algorithm selects an initial memory $\MemInit$ satisfying
$\Init_{\SafeReach}(\MemInit)\wedge\WinningMem(\MemInit)$ (line~7) and calls
$\textsc{ExtractStrategy}$, implemented using Skolem-function extraction in the
style of ~\cite{FarzanKincaid18}
\ifextended
(see Appendix~\ref{appendix:strategy-extraction}),
\else
(see the appendix of the extended version of the paper~\cite{extended_paper}),
\fi
to construct the policy and
memory-update function (line~8).
Concretely, the extraction procedure chooses Skolem functions for the
existential choices in the predecessor condition: $\Policy$ chooses an action
for each winning memory, and $\MemUpdate$ chooses a successor memory for each
enabled observation. By construction, these functions satisfy:
\begin{equation}\label{eq:extract-closure}
\forall m,o.\
\WinningMem(m)
\wedge
\Enabled_{\SafeReach}(m,\Policy(m),o)
\Rightarrow
\WinningMem(\MemUpdate(m,o))
\wedge
\Model_{\SafeReach}(m,\Policy(m),o,\MemUpdate(m,o)).
\end{equation}
Thus, $\MemInit$, $\Policy$, and $\MemUpdate$ collectively define the memory-based strategy
$\Strat$.

\begin{proposition}
\label{prop:solve-memory-correct}
If Algorithm~\ref{alg:solve-mem} returns $\mathit{Success}(\MemInit,\Policy,\MemUpdate)$,
  then $\Strat=\langle\Mem,\MemInit,\MemUpdate,\Policy\rangle$ is a winning strategy
  for $\game$.
  Conversely, if it returns $\mathit{Failure}$, then $\game$
  has no winning strategy that admits an epistemic certificate of the form
  $(\SafeReach,\Phi)$ with $\Phi\Rightarrow\Safe_{\SafeReach}$.
\end{proposition}

\subsection{Learning Memory Domains and Epistemic Support}
\label{sec:forward}

In this section, we turn our attention to the first phase of the algorithm for
inferring the memory domain and its epistemic support relation. As  summarized in Algorithm~\ref{alg:compute-safereach}, this phase first uses bounded
exploration to observe what uncertainty can arise in the game, and then
generalizes these observations into a symbolic memory representation.

The first step, implemented by \textsc{ExploreBeliefs}, explores the game up to
the current horizon $h$ using the most precise information available to the
controller: its \emph{full observable history}, consisting of the actions it has
taken and the observations it has received.
In this step, 
we symbolically compute a 
\emph{belief relation} $B(\ohist,s)$ that characterizes the states
$s$ that could result from some safe run consistent with observable history $\ohist$.  For any concrete history
$\mu$, instantiating $\ohist$ with $\mu$ yields a formula $B_\mu(s)$ over states;
we call this formula the \emph{belief} arising from $\mu$.

\begin{algorithm}[t]
  \caption{\textsc{InferMemory}$(\Init,\Model,\Safe,h)$}
  \label{alg:compute-safereach}
  \begin{algorithmic}[1]
    \State $\mathit{Beliefs} \gets
           \textsc{ExploreBeliefs}(\Init,\Model,\Safe,h)$
    \If{$\mathit{Beliefs} = \textsc{Failure}$}  \Return $(\bot, \_)$
    \EndIf
    \State $(M,\SafeReach) \gets
           \textsc{Abstract}(\mathit{Beliefs},\Safe)$
    \State \Return $(M,\SafeReach)$
  \end{algorithmic}
\end{algorithm}

In a sense, the full observable history can be viewed as the
most precise memory  possible, since it preserves all information that can be inferred from
the past. However, it is not a viable representation for an infinite-duration
strategy, because the history grows with the length of the play. Thus, we use 
these histories only as a temporary device for exposing the
uncertainty that can arise. The next step, implemented by \textsc{Abstract},
then generalizes the recurring symbolic structure in these history-indexed
beliefs into a fixed memory domain and an epistemic support relation. The rest of this section describes these two steps in turn.

\subsubsection{\bf Belief Exploration}

The first step of \textsc{InferMemory} explores the game up to the current
horizon and records the uncertainty induced by each \emph{observable history}, defined as follows:

\begin{definition}[Observable history]
A \emph{concrete observable history} is a finite sequence
$ 
  \mu = a_0 \cdot o_1  \cdots a_{k-1}\cdot o_k
$
of concrete controller actions and observations. We write $\epsilon$ for the
empty history and $\mu\cdot a\cdot o$ for the history obtained by extending
$\mu$ with action $a$ and observation $o$.

\end{definition}

Our belief exploration algorithm represents histories symbolically. A \emph{symbolic history}
$\ohist$ of length $k$ has the same shape as a concrete history, but its actions
and observations are variables. Each concrete
history $\mu$ of length $k$ is an instantiation of such a symbolic history. We write $\mathcal{H}_k$ for the histories of length~$k$ and $\mathcal{H}_{\leq h} = \cup_{k\leq h} \mathcal{H}_k$ for the set of all histories up to length~$h$.

We  associate each symbolic history with the set of states that  can result
from a safe run consistent with that history. We capture this through a \emph{belief relation}:

\begin{definition}[Belief]\label{def:belief}
Given a symbolic history $\ohist$, the belief relation $B_k(\ohist,s)$
characterizes the concrete states $s$ that can arise after some safe run
consistent with $\ohist$. It is defined inductively by:
\begin{equation}\label{eq:belief}
\begin{aligned}
  (1) \ \   B_0(\epsilon,s)
 \equiv \Init(s)\wedge\Safe(s) & \quad
  (2) \ \  B_{k+1}(\ohist\cdot a\cdot o,\,s')
    \equiv
    \Safe(s')\wedge \exists s.\,
      B_k(\ohist,s)\wedge\Model(s,a,s',o)
\end{aligned}
\end{equation}
\end{definition}

Thus, the initial belief consists of the safe initial states. The successor
belief keeps exactly those safe states $s'$ that can be reached from some state
in the previous belief by taking action $a$ and producing observation $o$.
In this sense, belief exploration is a symbolic version of the classical subset
construction~\cite{Reif84}, restricted to prefixes that have remained safe.

In addition to knowing the beliefs associated with a history, we also need to know which histories can still be extended safely
to a given horizon. To capture this, \textsc{ExploreBeliefs} computes a backward
winning predicate $\mathsf{Win}_k$ over symbolic histories of length $k \le h$:
\begin{equation}\label{eq:win}
\begin{aligned}
  \mathsf{Win}_h(\ohist)
    &\equiv \top \\
  \mathsf{Win}_k(\ohist)
    &\equiv
    \exists a.\ \forall s,s',o.\
      (B_k(\ohist,s)\wedge\Model(s,a,s',o)\bigr)
      \Rightarrow{}
      \bigl(\Safe(s')\wedge
      \mathsf{Win}_{k+1}(\ohist\cdot a\cdot o))
      \qquad\text{for } k<h
\end{aligned}
\end{equation}

Intuitively, $\mathsf{Win}_k(\ohist)$ says that the belief induced by history
$\ohist$ can be controlled safely for the remaining $h-k$ steps. That is, the controller
has an action such that every successor state is safe and every resulting
extended history is again winning. Hence, the controller can keep the game safe
up to horizon $h$ exactly when $\mathsf{Win}_0(\epsilon)$ holds.

\begin{algorithm}[t]
  \caption{\textsc{ExploreBeliefs}$(\Init,\Model,\Safe,h)$}
  \label{alg:explore-beliefs}
  \begin{algorithmic}[1]
    \State $B_0(\epsilon,s) \gets \Init(s)\wedge\Safe(s)$
    \For{$k = 1$ \textbf{to} $h$}
      \State $B_k(\ohist\cdot a\cdot o,\,s') \gets
        \Safe(s')\wedge\exists s.\ B_{k-1}(\ohist,s)\wedge\Model(s,a,s',o) \ | \ \ohist\in\mathcal{H}_{k-1}, a\in A, o \in O, s'\in S$
    \EndFor
    \State $\mathsf{Win}_h(\ohist) \gets \top$
    \For{$k = h-1$ \textbf{to} $0$}
      \State $\mathsf{Win}_k(\ohist) \gets
        \exists a.\ \forall s,s',o.\
        (B_k(\ohist,s)\wedge\Model(s,a,s',o))
        \Rightarrow (\Safe(s')\wedge\mathsf{Win}_{k+1}(\ohist\cdot a\cdot o))\ |\ \ohist\in \mathcal{H}_k$
    \EndFor
    \If{$\neg\,\mathsf{Win}_0(\epsilon)$} \Return \textsc{Failure}
    \EndIf
    \State \Return $[\,B_0\wedge\mathsf{Win}_0,\ldots,B_h\wedge\mathsf{Win}_h\,]$
  \end{algorithmic}
\end{algorithm}

Based on the forward belief recurrence in Eq.~\ref{eq:belief} and the backward
winning recurrence in Eq.~\ref{eq:win}, Algorithm~\ref{alg:explore-beliefs}
computes which beliefs are both reachable by safe prefixes and extendable to the
bounded horizon. Concretely, the
first loop computes the forward beliefs up to horizon $h$. The second loop works
backwards from the horizon to determine which histories are winning for the
remaining number of steps. If the empty history is not winning (i.e., 
$\neg\,\mathsf{Win}_0(\epsilon)$), then there  is no strategy that can
guarantee safety even for the bounded horizon $h$, so the procedure returns
\textsc{Failure}.
Otherwise, \textsc{ExploreBeliefs} returns only the parts of the beliefs from
which safety can still be guaranteed up to the horizon. In particular, for each
$k\le h$, the returned predicate is
$ 
  B_k(\ohist,s)\wedge\mathsf{Win}_k(\ohist)$, 
where the conjunct $\mathsf{Win}_k(\ohist)$ filters out histories that are safe
so far but cannot be extended safely to depth $h$.

\begin{theorem}
\label{thm:explore-beliefs-correctness}
\textsc{ExploreBeliefs}$(\Init,\Model,\Safe,h)$ provides the following guarantees:
\begin{enumerate}[leftmargin=*]
  \item \textbf{Safety.} If it returns $[B_0,\ldots,B_h]$, then we have $\forall k \le h. \  B_k(\ohist,s)\Rightarrow\Safe(s)$.
  \item \textbf{Refutation.} If it returns \textsc{Failure}, the
 original game has no winning strategy.
\end{enumerate}
\end{theorem}

\begin{example}
Consider the navigation game from Section~\ref{sec:overview}.
Initially, the controller knows only that the robot starts at $x=0$, with
altitude in $[0,6]$ and unknown bias:
\[
  B_0(\epsilon,s) \equiv
  s.x = 0 \wedge 0 \le s.y \le 6
  \wedge (s.b=1 \vee s.b=-1).
\]
Now consider the symbolic one-step history $\epsilon\cdot u_0\cdot o_1$ with
$o_1.x=1$. The corresponding belief $  B_1(\epsilon\cdot u_0\cdot o_1,s) $ is given by:
\[
\begin{aligned}
  -1 \le u_0 \le 1
  \wedge s.x=1
  \wedge s.z=o_1.z
  \wedge{}
  &\bigl(
    (s.b=1
     \wedge s.y=o_1.z-1
     \wedge 0 \le o_1.z-1-u_0 \le 6)
    \vee{}\\
  &\quad
    (s.b=-1
     \wedge s.y=o_1.z+1
     \wedge 0 \le o_1.z+1-u_0 \le 6)
  \bigr)
\end{aligned}
\]
The two disjuncts correspond to the two possible values of the hidden bias. The
additional bounds come from the initial altitude constraint: before applying
$u_0$, the altitude must have been in $[0,6]$.

Now consider one more symbolic step, with action $u_1$ and observation $o_2$
satisfying $o_2.x=2$. Since the beacon is no longer active at this position,
$o_2.z$ imposes no constraint on the altitude. Applying the forward recurrence
to the belief above yields the new belief $  B_2(\epsilon\cdot u_0\cdot o_1\cdot u_1\cdot o_2,s)$:
\[
\begin{aligned}
  s.x=2 \wedge -1 \leq u_0, u_1, \leq 1 \wedge
  &\bigl(
    (s.b=1 
     \wedge s.y=o_1.z+u_1-1
     \wedge 0 \le o_1.z-1-u_0 \le 6)
    \vee{}\\
  &\quad
    (s.b=-1
     \wedge s.y=o_1.z+u_1+1
     \wedge 0 \le o_1.z+1-u_0 \le 6)
  \bigr)
\end{aligned}
\]
\end{example}

\subsubsection{\bf Abstracting Beliefs into a Symbolic Memory Domain}
\label{sec:abstract-belief}

The goal of the \textsc{Abstract} procedure is to generalize the beliefs
encountered during bounded exploration into a fixed set of memory variables
whose values summarize  safety-relevant information. Our key
observation is that many history-indexed beliefs share a common formula schema
and differ only in the values of parameters derived from the history.
The abstraction phase aims to make these parameters explicit as memory variables.

We first formalize the notion of a \emph{memory domain} and then describe
our generalization procedure.

\begin{definition}[Symbolic memory domain]
A \emph{symbolic memory domain} $\Mem$ is a tuple of memory variables $\bar m$,
interpreted over some background theory. A memory value is a valuation of
$\bar m$. An \emph{epistemic support relation} for $\Mem$ is a formula
$\SafeReach(\bar m,s)$ relating memory values to concrete states.
\end{definition}
In other words, the memory variables specify the form of information retained by
the strategy, while the epistemic support $\SafeReach$ specifies what that
information actually means in terms of concrete states. The goal of
\textsc{Abstract} is to find memory variables $\bar m$ and an epistemic support
relation $\SafeReach(\bar m,s)$ such that each nonempty belief encountered during
\textsc{ExploreBeliefs} is represented by some valuation of $\bar m$. Formally,
for every  history $\ohist$ considered during belief exploration, we want:
\[
  \bigl(\exists s.\, B(\ohist,s)\bigr)
  \Rightarrow
  \exists \bar v.\,
  \forall s'.\;
  B(\ohist,s') \Leftrightarrow \SafeReach(\bar v,s') .
\]
This ensures that the memory valuation $\bar v$ represents exactly the same set
of possible concrete states as history $\ohist$, while still allowing the learned
support relation $\SafeReach(\bar m,s)$ to generalize to unseen game prefixes
through its use of symbolic memory variables $\bar m$.

\begin{algorithm}[t]
\caption{\textsc{Abstract}$(\mathit{Beliefs},\Safe)$}
\label{alg:abstract}
\begin{algorithmic}[1]
\State $B_{\mathit{hist}}(\ohist,s) \gets
  \bigvee_{i=0}^{h} \mathit{Beliefs}[i](\ohist,s) \ |\ \ohist\in\mathcal{H}_{\leq h},s\in S$
\State $\mathcal C \gets \emptyset$
\While{$\mathsf{Sat}\!\left(
  \exists \ohist,s_0.\;
  B_{\mathit{hist}}(\ohist,s_0)
  \wedge
  \neg
  \bigvee_{(S,T)\in\mathcal C}
  \exists \bar p.\;
  \forall s.\;
  B_{\mathit{hist}}(\ohist,s) \Leftrightarrow T(\bar p,s)
\right)$}
  \State $\mu \gets
    \mathsf{Model}\!\left(
      \exists \ohist,s_0.\;
      B_{\mathit{hist}}(\ohist,s_0)
      \wedge
      \neg
      \bigvee_{(S,T)\in\mathcal C}
      \exists \bar p.\;
      \forall s.\;
      B_{\mathit{hist}}(\ohist,s) \Leftrightarrow T(\bar p,s)
    \right)$
  \State $B_{\mathit{cex}}(s) \gets B_{\mathit{hist}}(\mu(\omega),s)$
  \State $\mathit{merged} \gets \mathit{false}$
  \ForAll{$(S,T) \in \mathcal C$}
    \State $T' \gets
      \textsc{SynthesizeTemplate}(S \cup \{B_{\mathit{cex}}\},\Safe)$
    \If{$T' \neq \textsc{Failure}$}
      \State $\mathcal C \gets
        (\mathcal C \setminus \{(S,T)\}) \cup
        \{(S \cup \{B_{\mathit{cex}}\},T')\}$;  $\mathit{merged} \gets \mathit{true}$; \textbf{break}
    \EndIf
  \EndFor
  \If{$\neg \mathit{merged}$}
    \State $T' \gets
      \textsc{SynthesizeTemplate}(\{B_{\mathit{cex}}\},\Safe)$; $\mathcal C \gets
      \mathcal C \cup \{(\{B_{\mathit{cex}}\},T')\}$
  \EndIf
\EndWhile
\State let $\mathcal C = \{(S_i,T_i)\}_{i=1}^k$
\State $\bar m \gets (c,\bar p)$, where $c\in\{1,\ldots,k\}$
\State $\SafeReach(\bar m,s) \gets
  \bigvee_{i=1}^{k} \left(c=i \wedge T_i(\bar p,s)\right)$
\State \Return $(\bar m,\SafeReach)$
\end{algorithmic}
\end{algorithm}

At a high level, \textsc{Abstract} achieves this goal 
by looking for beliefs with the
same symbolic shape. For example, two beliefs may be identical except that one contains the
constant $3$ where the other contains the constant $5$. The procedure groups
such beliefs together and replaces the differing constants with memory
variables, yielding a parameterized support predicate for each group.

Concretely, Algorithm~\ref{alg:abstract} implements this idea by maintaining a set
$\mathcal C$ of clusters. Each cluster $(S,T)\in\mathcal C$ consists of a finite
set $S$ of sample beliefs and a parameterized support predicate $T(\bar m,s)$
that represents those samples. Once all explored beliefs have been assigned to
clusters, the learned epistemic support relation $\SafeReach(\bar m,s)$ is
obtained by combining the predicates $T$ from all clusters.

In more detail, the first line of Algorithm~\ref{alg:abstract} combines the
beliefs discovered by \textsc{ExploreBeliefs} into a single 
formula $B_{\mathit{hist}}(\ohist,s)$, and the second line initializes the set
of clusters $\mathcal C$ to be empty. The main loop then repeatedly searches for
an explored history whose belief is not yet represented by any current cluster.
The satisfiability query in the loop guard asks whether there exists a history
$\ohist$ such that $B_{\mathit{hist}}(\ohist,s)$ is not equivalent to any
instantiation of any current cluster predicate $T$. If such a history exists,
the model provides a valuation $\mu$, and the algorithm extracts the uncovered
belief $ B_{\mathit{cex}}(s) $ as 
$  B_{\mathit{hist}}(\mu,s)$. 

Next, the algorithm  tries to incorporate $B_{\mathit{cex}}$ into one of the
existing clusters. For each cluster $(S,T)\in\mathcal C$, it calls
\textsc{SynthesizeTemplate} on the enlarged sample set
$S\cup\{B_{\mathit{cex}}\}$. If this call succeeds, the old cluster is replaced
by a new cluster whose sample set is $S\cup\{B_{\mathit{cex}}\}$ and whose
predicate is the newly synthesized predicate $T'$. If no existing cluster can be
generalized to include $B_{\mathit{cex}}$, the algorithm creates a fresh cluster
containing only this belief.

The  loop terminates only when the guard is unsatisfiable, meaning that every
belief is represented by some cluster predicate. The final line combines the
cluster predicates into the learned epistemic support relation
$\SafeReach(\bar m,s)$. If multiple clusters are produced, the memory domain
also includes a finite-valued variable $c$ indicating which cluster is being used. The remaining memory variables are  parameters of the clusters' support
predicates.

\begin{algorithm}[t]
\caption{\textsc{SynthesizeTemplate}$(S,\Safe)$}
\label{alg:synthesize-template}
\rev{
\begin{algorithmic}[1]
\Require Sample beliefs $S$, safety predicate $\Safe$
\Ensure Parameterized template $T$ or \textsc{Failure}
\State $B_0 \gets \textsc{Normalize}(\textsc{Choose}(S))$
\ForAll{parameterizations $U(\bar p,s)$ of $B_0$
        in increasing order of $|\bar p|$}
    \State $T(\bar p,s) \gets U(\bar p,s) \wedge \Safe(s)$
    \If{$\forall B\in S.\;
        \mathsf{Sat}\bigl(
        \exists \bar p.\ \forall s.\
        T(\bar p,s)\Leftrightarrow B(s)\bigr)$}
         \Return $T$
    \EndIf
\EndFor
\State \Return \textsc{Failure}
\end{algorithmic}
}
\end{algorithm}

\delrev{Given a finite set $S$ of sample beliefs,
\textsc{SynthesizeTemplate} searches for a parameterized predicate
$T(\bar p,s)$ by replacing selected constants and terms in the samples
with fresh parameters, which become memory variables. For example,
if two samples differ only in the constants appearing in corresponding
linear inequalities, the resulting template replaces those constants
with parameters. A candidate template is accepted only if it satisfies
the following conditions:}

\rev{Algorithm~\ref{alg:synthesize-template} describes the
\textsc{SynthesizeTemplate} procedure. Given a finite set $S$ of
sample beliefs, it first normalizes one of the beliefs and then
mines parameterized templates by replacing selected numerical
constants with symbolic parameters. Multiple occurrences may share
a parameter, allowing the template to preserve relationships
between different constraints. The procedure considers
these generalizations in increasing order of the number of
parameters, favoring simpler memory representations.
Each template is checked semantically against the sample beliefs
and the safety predicate. A template $T(\bar p,s)$ is accepted only if it satisfies
the following conditions:}

\begin{enumerate}[leftmargin=*]
  \item For every  belief $B$, there exists a valuation of the
  memory variables that represents $B$ exactly:
  \[
    \forall B \in S.\;
    \exists \bar p.\;
    \forall s.\;
    T(\bar p,s) \Leftrightarrow B(s)
  \]

  \item The template represents only safe states:
$ 
    \forall \bar p,s.\;
    T(\bar p,s) \Rightarrow \Safe(s)
$
\end{enumerate}

\delrev{The first condition ensures that $T$ is expressive enough to
represent all sample beliefs in the cluster, while the second ensures
that the generalization does not introduce unsafe states.}
\rev{The first condition ensures that generalization does not lose
information obtained during belief exploration, since each sampled
belief must be represented exactly by some parameter valuation.
At the same time, other parameter valuations may represent sets of
states not encountered during exploration, thereby allowing generalization. The second condition
ensures that these generalizations do not introduce unsafe states.
Our procedure enforces this condition by conjoining each mined
template with the original safety predicate $\Safe$.}

\begin{example}
Consider again the navigation game from Section~\ref{sec:overview}.
{For simplicity, we show only the constraints involving the
latent altitude and sensor bias, omitting the remaining state
constraints.}
After the robot has passed the beacon, different  histories  produce
beliefs with the same structure but different constants, such as:
\[
\begin{array}{rcl}
  B_1(s) &\equiv&
  (s.b=1 \wedge s.y=3/2-1)
  \vee
  (s.b=-1 \wedge s.y=3/2+1), \\[0.6ex]
  B_2(s) &\equiv&
  (s.b=1 \wedge s.y=5/2-1)
  \vee
  (s.b=-1 \wedge s.y=5/2+1).
\end{array}
\]
These beliefs have the same symbolic shape; they differ only in the value
$3/2$ versus $5/2$. \textsc{SynthesizeTemplate} generalizes them by replacing such constants with a memory variable $w$, yielding:
\[
  T(w,s) \equiv
  (s.b=1 \wedge s.y=w-1)
  \vee
  (s.b=-1 \wedge s.y=w+1).
\]
This template represents both samples exactly (e.g., we have $  \forall s.\; B_1(s) \Leftrightarrow T(3/2,s)$).
\end{example}

\subsection{Properties of the Algorithm}

\rev{Having described the individual components of our synthesis
procedure, we now establish the guarantees of \textsc{FindWinning}
as a whole. We first prove its soundness, then discuss its
limitations and establish conditions under which it is
guaranteed to succeed.}

\begin{theorem}[Soundness of \textsc{FindWinning}]
\label{thm:findwinning-sound}
Algorithm~\ref{alg:solve-po-game} satisfies the following guarantees:
\begin{enumerate}[leftmargin=*]
\item \textbf{Strategy soundness.}
If \textsc{FindWinning} returns a strategy
$\Strat=\langle \Mem,\MemInit,\MemUpdate,\Policy\rangle$, then $\Strat$ is indeed a winning
strategy for $\game$.

\item \textbf{Refutational soundness.}
If \textsc{FindWinning} returns \textsc{Fail}, then $\game$ has no
winning strategy.
\end{enumerate}
\end{theorem}

As expected in the infinite-state setting, Algorithm~\ref{alg:solve-po-game}  does not have termination guarantees. This limitation is inherent, as the problem is undecidable even for infinite-state perfect-information safety games. Our algorithm is  therefore \emph{not} a full decision procedure and exhibits two ways in which incompleteness can arise. First, even when the game is unrealizable, there need not be a uniform finite
horizon by which every strategy is forced to violate safety. Thus, bounded
exploration may fail to produce a refutation.  Second, even when a winning strategy
exists, the memory abstractions inferred from bounded histories may fail to
admit an epistemic certificate, in which case the algorithm may also continue
refining without success. 

\rev{Despite these limitations, we can establish a conditional
success guarantee that identifies when the overall
synthesis procedure succeeds.}

\begin{theorem}[Conditional Success]
\label{thm:conditional-success}
\rev{
Suppose that, at some iteration, \textsc{InferMemory} returns
a memory domain $M$ and an epistemic support relation $\Psi$
such that there exists a strategy $\mathcal S$ admitting
an epistemic certificate $(\Psi,\Phi)$ with
$\Phi(m) \Rightarrow \phi_\Psi(m)$.
If the fixpoint computation in \textsc{SolveMemoryGame}
terminates, then \textsc{FindWinning} returns a winning
strategy at that iteration.
}
\end{theorem}

\section{Implementation}
\label{sec:impl}

We have implemented our approach in a new prototype tool called \tool{},\footnote{{We used Claude Code to assist with the development of \tool{}.}}
which is written in
Python using the Z3 SMT solver~\cite{z3}.  The input to \tool{} is a
symbolic game model and safety specification, expressed in SMT-LIB format~\cite{smtlib}.
The current implementation targets formulas in linear arithmetic, which is the
background theory used by all benchmarks in our evaluation.
\tool{} represents the synthesized policy as a finite set of guarded cases, where each case specifies an
action to take when the current memory value satisfies a corresponding guard.
Similarly, the memory-update function is represented as guarded cases that
specify how to compute the next memory value.

\bfpara{Solver interface.}
\tool{} uses the  Z3  SMT solver for  satisfiability, validity, and projection queries. For example, during
belief exploration, \tool{} relies on  Z3 for computing symbolic post- and pre-conditions, and, during generalization, for finding uncovered beliefs and synthesizing templates. Similarly, \tool{}  uses Z3 in the memory game solving phase for computing the winning memory region and extracting a winning strategy.
Projection is implemented using Z3's quantifier
elimination support.  While  \tool{} could also be
extended with  \emph{approximate} projection operators,  our current prototype only supports exact quantifier elimination. Thus, in its current form, \tool{} is primarily intended for solving games whose underlying theory is linear arithmetic (LRA or LIA).

\bfpara{Optimizing belief exploration.}
\tool{} implements two optimizations to make belief exploration more efficient.
Algorithm~\ref{alg:explore-beliefs} prioritizes clarity of exposition, so it
presents belief exploration in its simplest form, but our implementation is optimized to avoid unnecessary solver work.

First, to avoid exploring
unnecessary beliefs, \tool{} uses an incremental variant of belief exploration.
Rather than treating the forward belief computation and the winning check as two
completely separate phases, \tool{} uses the winning check to prune
the search space.  Once a history is known to be unable to
support safety up to the current horizon, later iterations do not expand it
further
\ifextended
(see Appendix~\ref{appendix:incremental-belief-exploration}).
\else
(see the appendix of the extended version of the paper~\cite{extended_paper}).
\fi
Second, because quantifier elimination introduces large formulas with many
redundant constraints,  \tool{} applies a simplification
pass after each quantifier-elimination step. This pass uses Z3's simplifier
together with constraint minimization~\cite{DilligSAS2010Simplify} to remove
redundant constraints.

\newcommand{\rot}[1]{\rotatebox{90}{\hspace{1pt}#1}}

\section{Evaluation}\label{sec:eval}
In this section, we describe an  evaluation of \tool{} designed to
answer the following  questions:
\begin{enumerate}[label=\textbf{RQ\arabic*.}, leftmargin=*]
  \item Can \tool{} synthesize winning strategies for partially
  observable safety games with \emph{both} infinite state spaces and infinite
action alphabets?

  \item On benchmarks with \emph{finite} action alphabets, how does
  \tool{} compare to prior approaches for solving partially observable safety
  games?

  \item How do the key algorithmic ingredients of \tool{}  affect its
  effectiveness?
    \item Do the learned memory abstractions adapt to the safety-relevant
uncertainty induced by the observation model?
\end{enumerate}

All experiments are conducted on a machine with an Intel Xeon E5-2640 v3 CPU with a per-run memory limit of
\SI{64}{GB} and a per-run time limit of 300\,s.

\subsection{Evaluation on Games with Infinite State and Action Spaces}
\label{sec:eval-rq1}

To answer RQ1, we evaluate \tool{} on symbolic games with infinite
state spaces and real-valued controller actions.
\delrev{In total, we consider 30 benchmarks generalized from prior
work~\cite{AtkinsonCarbin2020PartialObservability,
agarwal_nsdi24_belief_cc,DimitrovaThesis}.
Our benchmark suite spans UAV control, robot navigation,
lander descent control, and congestion control.
Across these domains, the controller must act under incomplete
information while choosing continuous control values, such as
altitude corrections, movement commands, engine-shutoff decisions,
and sending-rate updates.}
\rev{Since there is no established benchmark suite for this setting,
we systematically generalize models from prior
work~\cite{AtkinsonCarbin2020PartialObservability,
agarwal_nsdi24_belief_cc,DimitrovaThesis}
along dimensions that affect observability, control, and safety.
This yields a suite of 30 benchmarks, summarized in Table~\ref{tab:benchmark-models}, spanning four application
domains.}
\delrev{The benchmark suite includes both realizable and unrealizable
instances, allowing us to evaluate both strategy synthesis
and unrealizability detection.}
\rev{The suite includes both realizable and unrealizable instances,
allowing us to evaluate both strategy synthesis and unrealizability
detection.}

\begin{table}[t]
\vspace{-0.17in}
\caption{\rev{Benchmark characteristics and variations used in RQ1.}}
\vspace{-0.17in}
\centering
\small
\setlength{\tabcolsep}{5pt}
\renewcommand{\arraystretch}{1.12}
\begingroup
\ifshowrevisions\color{blue}\fi

\begin{tabularx}{\linewidth}{@{}>{\bfseries}lYYY@{}}
\toprule
Family & Partial observability & Safety objective & Variations \\
\midrule

UAV
& Noisy position sensing; unknown disturbances
& Remain within safe flight region
& 1D--3D; sensing and disturbance uncertainty; safety margins \\
\addlinespace[0.6em]

Robot
& Missing, imprecise, or delayed localization
& Avoid obstacles and satisfy target constraint
& %
Sensing accuracy; movement choices \\
\addlinespace[0.6em]

Lander
& Unreliable touchdown signals
& Avoid premature or delayed engine shutdown
& Sensor-error margins \\
\addlinespace[0.6em]

CCA
& Unknown bandwidth and buffer state; delayed feedback
& Bound packet loss and rate overshoot; require progress below target
& Buffer size; feedback delay and jitter; control granularity \\

\bottomrule
\end{tabularx}
\endgroup
\label{tab:benchmark-models}
\vspace{-0.2in}
\end{table}

\bfpara{Benchmarks.}
\delrev{Table~\ref{tab:rq1} summarizes the benchmark suite used in
our RQ1 evaluation. We organize these benchmarks into four families:}
\rev{We describe the four benchmark families from Table~\ref{tab:benchmark-models} in more detail,
including their origins and the challenges they pose for synthesis.}

\begin{enumerate}[leftmargin=*]

\item \textbf{UAV control.}
These benchmarks are continuous-state synthesis generalizations of
the UAV altitude-control example from Atkinson and Carbin's work
on belief programming~\cite{AtkinsonCarbin2020PartialObservability}.
\delrev{Their example verifies a hand-written controller for maintaining
altitude under bounded wind disturbance and noisy observations.
We adapt this example to the synthesis setting and generalize
the original one-dimensional example to infinite-state 2D and 3D
variants. We also construct unrealizable variants obtained by
tightening the safety envelope or increasing uncertainty.}
\rev{Whereas their work verifies a hand-written controller, our
goal is to synthesize control actions that keep the UAV within
a designated safe flight region despite uncertain motion.
Partial observability arises because the controller receives
noisy position measurements rather than the UAV's true position.
Both sensor errors and motion disturbances range over bounded
continuous intervals. We extend the original one-dimensional model to 2D and 3D
and vary sensing uncertainty, disturbance bounds, and safety
margins. Higher-dimensional variants increase the complexity
of the control problem, while greater uncertainty or tighter
safety margins make it more difficult to guarantee safety, 
also yielding unrealizable instances.}

\item \textbf{Robot navigation.}
These benchmarks generalize the robot motion-planning examples
from Dimitrova's thesis~\cite{DimitrovaThesis}
from finite-action navigation to infinite-action control.
\delrev{The original examples involve a robot with real-valued
two-dimensional position that must reach a target region while
avoiding walls under partial observability.
We adapt them into a parameterized family of partially observable
safety games in which the controller may choose continuous
movement directions and/or step sizes.
We scale the family by enlarging the workspace, adding obstacles,
reducing sensor accuracy, and varying the actuation model.}
\rev{The synthesis task is to choose movements that allow a robot
with real-valued position to reach a target region while avoiding
obstacles, using imprecise or delayed position feedback rather
than its true position.
We generalize the original models by allowing continuous
step sizes, and vary the 
sensing accuracy and actuation model.
These variations capture differences in %
localization accuracy and robot maneuverability, both of which
affect the ability to navigate safely under uncertainty.}

\item \textbf{Lander descent control.}
These benchmarks generalize the Mars Polar Lander case study
of~\cite{AtkinsonCarbin2020PartialObservability} from verification
of a hand-written controller to synthesis of a shutdown policy.
The controller must decide when to shut down the engine during
descent.
\delrev{In particular, shutting down too early can cause a crash,
while shutting down too late violates the intended landing protocol.
Partial observability arises because faulty touchdown sensors
make individual readings inconclusive.
We vary the sensor-error margin to obtain both realizable and
unrealizable instances.}
\rev{Because touchdown sensor readings may be unreliable, the
controller must make this decision without knowing for certain
whether the lander has touched down. Shutting down too early
can cause a crash, while shutting down too late violates the
landing protocol.
We vary the sensor-error margin to model different levels of
uncertainty. Larger margins make it harder
to determine when shutdown is safe and can render the
specification unrealizable.}

\item \textbf{Congestion control.}
These benchmarks target the congestion-control setting studied by
Agarwal et al.~\cite{agarwal_nsdi24_belief_cc}.
\delrev{The controller must choose sending rates using indirect
network feedback while satisfying safety objectives such as
bounded packet loss and, in some instances, convergence to the
available bandwidth. The benchmarks vary the control granularity,
buffer assumptions, and the presence of jitter, including RTT-level
and global-timestep control, zero-buffer and finite-buffer
environments, and variants with or without jitter.}
\rev{The synthesis task is to choose sending rates that bound packet
loss and overshoot without directly observing the available
bandwidth or buffer state. We adopt the safety-based formulation
of congestion-control objectives from ~\cite{agarwal_nsdi24_belief_cc}, but unlike
their approach, which relies on manually constructed belief sets,
invariants, and controller templates, our method automatically
synthesizes both the memory abstraction and control policy.
We vary buffer capacity, feedback delay and jitter, and control
granularity. These variations affect how much excess traffic
the network can tolerate and how quickly the sender can react
to congestion.}

\end{enumerate}

\rev{
These benchmarks collectively cover a range of partially
observable safety synthesis problems across different
application domains. In particular, they capture distinct
sources of partial observability, including noisy measurements,
 missing localization information, and
delayed feedback. By systematically varying the observation
models, control choices, and safety requirements, we evaluate
\tool\ under different forms of uncertainty and levels of
synthesis complexity. The resulting suite includes both
realizable and unrealizable instances, allowing us to evaluate
the ability of \tool\ to synthesize winning strategies as well
as prove unrealizability.
}

\delrev{Although the benchmark domains are drawn from prior work,
we note that the corresponding tools do not provide a suitable
basis of comparison against \tool. Specifically,
~\cite{AtkinsonCarbin2020PartialObservability} provides a
programming model and verification logic for
\emph{manually-written controllers}. The congestion-control
work~\cite{agarwal_nsdi24_belief_cc} is also not directly
comparable due to its domain-specific nature that exploits
network-specific belief computations and templates.
The closest general-purpose approach is the CEGAR-based
method of Dimitrova and
Finkbeiner~\cite{DimitrovaFinkbeinerFSTTCS08,DimitrovaThesis},
but it assumes a finite action alphabet.}

\begin{table}[t]
\vspace{-0.17in}
\caption{RQ1 results on benchmarks. All reported times are in seconds.}
\vspace{-0.17in}
\centering
\small
\setlength{\tabcolsep}{5pt}
\renewcommand{\arraystretch}{1.08}
\begin{tabular}{l r r r r @{\hspace{8pt}} r r r}
\toprule
\multirow{2}{*}{\textbf{Family}} &
\multirow{2}{*}{\textbf{\#Inst}} &
\multirow{2}{*}{\textbf{\#Sat}} &
\multirow{2}{*}{\textbf{\#Unsat}} &
\multirow{2}{*}{\textbf{\#Solved}} &
\multicolumn{3}{c}{\textbf{{Solving time (s)}}} \\
\cmidrule(lr){6-8}
& & & & &
\textbf{Mean} & \textbf{Median} & \textbf{Max} \\
\midrule
UAV   & 6 & 3 & 3 & 6 (100\%) & 77.8 & 20.2  & 200.8 \\
Robot & 10  & 8  & 2 & 10 (100\%)  & 43.6 & 26.2 & 134.1 \\
{Lander} & 4  & 2  & 2 & 4 (100\%)   & 61.2 &  60.4 & 100.4  \\
CCA   & 10 & 7  & 3 & 9 (90\%)   & 32.3 & 3.8  & 218.4 \\
\midrule
\textbf{Overall} & \textbf{30} & \textbf{20} & \textbf{10} & \textbf{29 (97\%)} &
\textbf{49.4} & \textbf{15.6} & \textbf{218.4} \\
\bottomrule
\end{tabular}
\label{tab:rq1}
\vspace{-0.2in}
\end{table}

\bfpara{Results overview.}
Table~\ref{tab:rq1} summarizes the key results.
\rev{Our suite contains 20 realizable and 10 unrealizable
instances, allowing us to evaluate both strategy synthesis
and unrealizability detection.}
Overall, \tool{} solves 29 out of 30 instances (97\%),
with the only failure being in the CCA category.
\rev{In particular, \tool\ solves all instances in the UAV,
Robot, and Lander families, and 9 out of 10 CCA instances.}
\delrev{In terms of running time, the overall median is 15.6s,
indicating that most instances complete quickly.
The mean, on the other hand, is substantially larger
than the median (49.4s vs.\ 15.6s), with maximum
solving time reaching 218.4s. This indicates that
a small number of long-running instances dominate
the tail of the distribution.}
\rev{Among the solved instances, the median solving time
is 15.6s, although the mean is substantially higher
at 49.4s. Interestingly, the CCA family has the
lowest median solving time (3.8s), but also contains
the longest successful run (218.4s) and the only
unsolved instance. Thus, even within the same
application domain, the difficulty of synthesis
can vary substantially. The sole failure occurs
during quantifier elimination in the memory-game
fixpoint computation, highlighting symbolic game
solving as an important remaining scalability
bottleneck.}

\bfpara{Memory abstraction and strategy complexity.}
\rev{We also examine the complexity of the memory abstractions
and strategies synthesized by \tool.
Table~\ref{tab:memory-characteristics} summarizes these results.
Across all synthesized strategies, the learned abstractions
use a median of 4 templates and 2 memory variables, with
epistemic support relations containing a median of 79 AST nodes.
The number of templates and memory variables captures
different aspects of abstraction complexity.
For example, UAV uses only 2 templates but requires more
memory variables to represent multiple intervals.
In contrast, Robot uses more templates (median 5) but
fewer memory variables (median 2), reflecting a greater
variety of belief structures.
The complexity of the support relations also varies
considerably, from 33 AST nodes for UAV to 164 for Lander.
Interestingly, suitable abstractions are typically discovered
at relatively shallow exploration depths (median 3),
although Robot and CCA require up to 6 steps.
Finally, memory-update functions tend to be more complex
than control policies, with a median of 16 versus 6
guarded cases. This suggests that maintaining the
controller's knowledge can be more involved than deciding
how to act on it.}

\begin{table}[t]
\vspace{-0.17in}
\caption{Characteristics of learned memory abstractions and
synthesized strategies (median/maximum over successful instances).
Exploration depth is the maximum length of execution prefixes used to infer the successful memory abstraction.
The number of templates measures how many parameterized belief structures are needed in the learned epistemic support. Memory
variables exclude the current observation; support size is measured in AST nodes; policy and update sizes
are measured in guarded cases.}
\vspace{-0.17in}
\centering
\footnotesize
\setlength{\tabcolsep}{4pt}
\renewcommand{\arraystretch}{1.15}

\begingroup
\ifshowrevisions\color{blue}\fi
\begin{tabular*}{\linewidth}{@{\extracolsep{\fill}}lcccccc@{}}
\toprule
& \multicolumn{4}{c}{Memory abstraction}
& \multicolumn{2}{c}{Synthesized strategy} \\
\cmidrule(lr){2-5}
\cmidrule(l){6-7}
Family
& \shortstack{Exploration depth\\(med/max)}
& \shortstack{Templates\\(med/max)}
& \shortstack{Memory vars\\(med/max)}
& \shortstack{Support size\\(med/max)}
& \shortstack{Policy cases\\(med/max)}
& \shortstack{Update cases\\(med/max)} \\
\midrule
UAV     & 1 / 1 & 2 / 2 & 5 / 7 & 33 / 44   & 9 / 18   & 25 / 63 \\
Robot   & 4 / 6 & 5 / 6 & 2 / 2 & 80 / 91   & 7.5 / 11 & 11 / 22 \\
Lander  & 2 / 2 & 4 / 4 & 4 / 4 & 164 / 166 & 2 / 2    & 30 / 33 \\
CCA     & 3 / 6 & 3 / 4 & 3 / 4 & 62 / 115  & 3.5 / 6  & 15.5 / 32 \\
\midrule
Overall & 3 / 6 & 4 / 6 & 2 / 7 & 79 / 166  & 6 / 18   & 16 / 63 \\
\bottomrule
\end{tabular*}
\label{tab:memory-characteristics}
\endgroup
\vspace{-0.20in}
\end{table}

\ \\
\idiotbox{RQ1}{
\tool{} synthesizes winning strategies for partially observable safety games with infinite state and action spaces, solving {$29$} out of $30$ benchmarks ({$97\%$}) with a median solving time of {15.6s.} 
}

\vspace{0.1in}

\subsection{Comparison with Prior Work on Finite Action Spaces}
\label{sec:eval-rq2}

To answer RQ2, we compare \tool{} against SPIAR, the
incomplete-information baseline of Dimitrova and
Finkbeiner~\cite{DimitrovaFinkbeinerFSTTCS08}.
We use the implementation of SPIAR provided to us by the
authors and restrict the comparison to benchmarks where
both approaches are applicable, namely infinite-state
games with a \emph{finite} action alphabet.
\delrev{Since our goal is to compare the scalability of the
approaches as we scale the size of the action alphabet,
we perform this evaluation on three representative
benchmark families from~\cite{DimitrovaThesis} that
admit a meaningful ``action scaling'' knob.}
\rev{In this experiment, we focus on action-alphabet size because the key motivation for our approach is handling
large or infinite action spaces. To this end, we consider the following three
benchmark families from~\cite{DimitrovaThesis} in
which the available controller choices can be
systematically increased:}

\begin{itemize}[leftmargin=*]

\item \textbf{Parts.}
\rev{This benchmark models a robotic sorting system that 
routes parts to  bins based on their type.
The controller has only partial information about each part,
such as its size and weight, and must choose a destination
without directly observing its true type. The objective is
to ensure that every part is sorted correctly.
We scale the action alphabet by increasing the number of
part types and corresponding destinations. This gives the
controller more sorting choices while also increasing the
number of distinctions it must make from incomplete sensor
information. Thus, the experiment tests how the approaches
scale as the sorting task becomes more complex.}

\item \textbf{Bakery.}
\delrev{This is a program synthesis instance modeled after
Lamport's Bakery protocol. Process $A$ is fixed, while
process $B$ must be completed by choosing how to update
$\mathit{ticketB}$ and when to enter the critical section.
Here, increasing the action set corresponds to enlarging
the finite library of candidate update/guard choices.}
\rev{This benchmark considers completing a partially specified
implementation of Lamport's Bakery mutual-exclusion protocol.
One process follows the original protocol, while the
synthesizer must determine how the other process assigns
tickets and decides when to enter its critical section.
The synthesized process can observe its own state and
ticket, but has only partial information about the other
process. The goal is to ensure mutual exclusion and
bounded waiting.
We scale the action alphabet by increasing the number of
candidate ticket assignments available
to the synthesizer. This gives the synthesizer greater
freedom in completing the protocol and tests how the
approaches scale with the number of possible implementation
choices.}

\item \textbf{Robot.}
\rev{This benchmark considers synthesizing a controller for
a robot navigating between regions separated by walls.
The robot must reach a target region while avoiding
collisions, despite having incomplete information about
its position. The original model allows the controller
to choose among four movement directions.
We scale the action alphabet by introducing additional
horizontal and vertical displacement choices, giving
the controller finer control over its movements.
This provides increasingly precise approximations of
continuous control and tests how the approaches scale
as the number of available movement commands grows.}

\end{itemize}

\delrev{
The second benchmark family, \textsc{Parts}, models a
robotic sorting controller that routes items of
different types into bins. For this family, increasing
the action set corresponds to increasing the number
of item types (and thus sorting destinations).

Finally, \textsc{Robot} models a robot with real-valued
two-dimensional position that must reach a target region
while avoiding walls under partial observability.
The controller chooses from a finite set of movement
commands, each corresponding to a discrete choice of
horizontal and vertical displacement.
For this family, increasing the action set corresponds
to increasing the number of available $x$- and
$y$-displacements, thereby giving the controller a
finer approximation of the underlying continuous
action space.
}

\begin{figure}[t]
\vspace{-0.2in}
  \centering
\includegraphics[width=\linewidth]{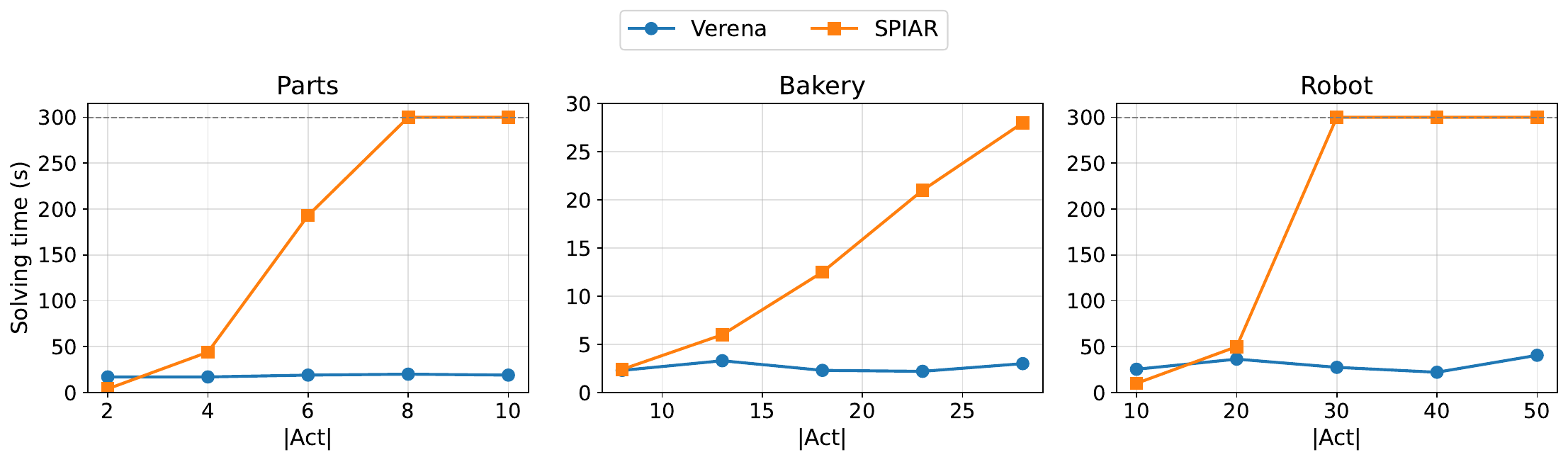}
\vspace{-0.3in}
  \caption{RQ2 experiment: Comparison against {\sc SPIAR}~\cite{DimitrovaFinkbeinerFSTTCS08} on three benchmark families with finite action spaces. The $x$-axis is the size of the action alphabet.}
  \label{fig:rq2}
\end{figure}

The results of this evaluation are shown in
Figure~\ref{fig:rq2}, where the $x$-axis shows the
size of the action alphabet and the $y$-axis is
end-to-end synthesis time, with the dashed line
at 300s showing the timeout threshold.
\delrev{Across all three benchmark families, \tool\
scales substantially better: Its runtime grows only
modestly as we increase the action alphabet and
remains well below the 300s timeout bound for all
instances. In contrast, the baseline (SPIAR)
degrades rapidly, exceeding the 300s timeout on
the Robot and Parts configurations, while still
rising to roughly 30s on the largest Bakery instance.}
\rev{Across all three benchmark families, \tool\
scales substantially better as the action alphabet
grows. Its runtime increases only modestly over
the tested configurations, remaining well below
the 300s timeout. In contrast, SPIAR's runtime
increases much more sharply, exceeding the timeout
on the larger Robot and Parts instances and reaching
roughly 30s on the largest Bakery instance.}

\delrev{This scaling gap can be attributed to how these two
methods deal with the action alphabet.
SPIAR~\cite{DimitrovaFinkbeinerFSTTCS08} relies on
explicit refinement of an abstract game and must
account explicitly for each available action.
Thus, enlarging the action alphabet directly
increases the amount of work it must perform.
In contrast, \tool\ reasons about actions symbolically
and builds a compact, belief-centric representation
of the strategy space. As a result, \tool\ scales
better as we increase the space of available actions.}
\rev{The difference is consistent with how the two
approaches handle controller actions.
SPIAR~\cite{DimitrovaFinkbeinerFSTTCS08} explicitly
represents the available actions in its abstract game,
so increasing the action alphabet increases the size
of the game that must be analyzed.
In contrast, \tool\ represents controller actions
symbolically and synthesizes them by solving logical
constraints, without enumerating individual actions.
Consequently, its runtime does not grow proportionally
with the number of available actions, although it
still depends on the complexity of the resulting
formulas and the underlying solver.}

\ \\
\idiotbox{RQ2}{
On benchmarks with finite action alphabets, \tool{} scales substantially better than \textsc{SPIAR}, remaining well below the 300\,s timeout as the action space grows, while \textsc{SPIAR} times out on several instances.
}

\vspace{0.1in}

\subsection{Ablation Studies}\label{sec:ablation}

To answer RQ3, we evaluate which parts of \tool{} are necessary for solving the
benchmarks. Recall that \tool{} proceeds in two conceptual phases. First, it
learns a symbolic memory abstraction together with an epistemic support relation
that explains what concrete states each memory value may represent. Second, it
uses this support relation to construct and solve the induced memory game,
thereby identifying the memory values from which safety can be maintained
indefinitely. Thus, for this  evaluation, we consider  ablations that remove or simplify different aspects of this
pipeline, as summarized in Table~\ref{tab:ablation-summary} and explained below:

\begin{table}[t]
\vspace{-0.17in}
\caption{Summary of ablations used for RQ3.}
\vspace{-0.17in}
\centering
\small
\setlength{\tabcolsep}{5pt}
\renewcommand{\arraystretch}{1.12}
\begin{tabular}{l l l}
\toprule
\textbf{Ablation} & \textbf{Change to \tool{}} & \textbf{Purpose} \\
\midrule
\textsc{TrivialSupport}
& use $\Psi_{\mathsf{triv}}(m,s)\equiv \Safe(s)$
& remove epistemic information \\

\textsc{SupportAsWin}
& use $\exists s.\,\Psi(m,s)$ as the winning region
& remove winning-region computation \\

\textsc{SuffixMem($k$)}
& store only the last $k$ observations
& remove learned symbolic memory \\
\bottomrule
\end{tabular}
\vspace{-0.20in}
\label{tab:ablation-summary}
\end{table}

\bfpara{\textsc{TrivialSupport}.}
This ablation removes the learned epistemic support relation. Concretely, it
uses the trivial support $
  \SafeReach_{\mathsf{triv}}(m,s) \equiv \Safe(s)$,
which records only that the concrete state is safe and forgets any connection
between memory values and the hidden state. Under this support, all memory
values have the same semantic interpretation, namely that they represent the set of safe
states. Hence,  this ablation tests whether the induced-game solver can recover a
winning strategy without the epistemic information learned by \tool{}.

\bfpara{\textsc{SupportAsWin}.}
This ablation uses the same symbolic memory and support relation
$\SafeReach$ as \tool{}, but treats every memory value with nonempty support
as part of the winning
region. The ablation then attempts to synthesize $(\Policy,\MemUpdate)$ from
this candidate region; if extraction fails, it increases the exploration horizon
and repeats. Hence, this ablation tests whether learned support is sufficient on its own, or
whether the backward fixpoint over the induced memory game is needed to identify
which supported memories are actually winning.

\bfpara{\textsc{SuffixMem($k$)}.}
This ablation assesses the importance of learning a problem-specific memory
domain. In particular, \textsc{SuffixMem($k$)} replaces the learned symbolic
memory abstraction with a fixed history-based memory: at step $i$, the memory
stores the suffix of the last $k$ observations, and the memory update is the
corresponding shift update. The rest of the pipeline remains unchanged. This
ablation tests whether a generic bounded observation history is sufficient.

\begin{table}[t]
\vspace{-0.17in}
\caption{Ablation results for RQ3.}
\vspace{-0.17in}
\centering
\small
\setlength{\tabcolsep}{3pt}
\renewcommand{\arraystretch}{1.1}
\begin{adjustbox}{max width=\linewidth}
\begin{tabular}{c c r r r r r r}
\toprule
\textbf{\shortstack{Benchmark\\Family}} &
\textbf{\shortstack{Num\\Instances}} &
\textbf{\shortstack{\tool{}\\Original}} &
\textbf{\shortstack{\textsc{Trivial}\\\textsc{Support}}} &
\textbf{\shortstack{\textsc{Support}\\\textsc{AsWin}}} &
\textbf{\shortstack{\textsc{Suffix}\\\textsc{Mem(1)}}} &
\textbf{\shortstack{\textsc{Suffix}\\\textsc{Mem}(2)}} &
\textbf{\shortstack{\textsc{Suffix}\\\textsc{Mem}(3)}} \\
\midrule
UAV           & \textsc{6}  & \textsc{6 (100\%)}  & \textsc{0 (0\%)}  & \textsc{3 (50\%)} & \textsc{5 (83\%)}  & \textsc{5 (83\%)}   & \textsc{5 (83\%)} \\
Robot         & \textsc{10} & \textsc{10 (100\%)} & \textsc{0 (0\%)}  & \textsc{2 (20\%)} & \textsc{2 (20\%)}  & \textsc{2 (20\%)}   & \textsc{2 (20\%)} \\
Lander        & \textsc{4}  & \textsc{4 (100\%)}  & \textsc{0 (0\%)}  & \textsc{2 (50\%)} & \textsc{2 (50\%)}  & \textsc{2 (50\%)}   & \textsc{2 (50\%)} \\
CCA           & \textsc{10} & \textsc{9 (90\%)}   & \textsc{1 (10\%)} & \textsc{3 (30\%)} & \textsc{3 (30\%)}  & \textsc{3 (30\%)}   & \textsc{3 (30\%)} \\
Finite Action & \textsc{15} & \textsc{15 (100\%)} & \textsc{5 (33\%)} & \textsc{5 (33\%)} & \textsc{10 (67\%)} & \textsc{15 (100\%)} & \textsc{15 (100\%)} \\
\midrule
\textbf{Total} & \textbf{45} & \textbf{44 (98\%)} & \textbf{6 (13\%)} & \textbf{15 (33\%)} & \textbf{22 (49\%)} & \textbf{27 (60\%)} & \textbf{27 (60\%)} \\
\bottomrule
\end{tabular}
\end{adjustbox}
\label{tab:ablation}
\vspace{-0.20in}
\end{table}

\bfpara{Results.} The results of this ablation study are summarized in Table~\ref{tab:ablation} and cover all benchmarks considered  in Sections~\ref{sec:eval-rq1} and ~\ref{sec:eval-rq2}.  Overall, the full \tool{} system solves  {44 (98\%)} of the  45 benchmarks. Among the three ablations, \textsc{TrivialSupport} solves the fewest, succeeding on only {13\%} of the benchmarks. This result illustrates that having a strong epistemic support is crucial. The \textsc{SupportAsWin} ablation does better, solving {33\%} of the benchmarks.
Finally, the fixed memory abstractions perform the best, achieving up to {60\%} accuracy for $ k \in \{2 , 3\}$. Notably, increasing $k$ from 1 to 2 allows solving an additional {$11\%$} of the benchmarks, but increasing $k$ further does not yield benefits. Overall, the significant difference between the full \tool{} system and its ablations demonstrates that all of our algorithmic choices are crucial for the efficacy of the approach. 

\ \\
\idiotbox{RQ3}{
Ablations show that our key algorithmic ingredients are necessary: \tool{} solves {$44/45$} benchmarks ({98\%}), while
removing key components drops performance sharply (down to {13--60}\% solved).
}

\vspace{0.1in}

\subsection{Qualitative Analysis of Learned Memory Abstractions and Strategy}
\label{sec:eval-rq4}
To answer RQ4, we inspect strategies synthesized by \tool{} on two closely
related congestion-control benchmarks. The goal of this case study is to show
that small changes in the observation model can change what the controller must
remember, even when the safety objective is the same. 

\bfpara{The set-up.}
Both benchmarks have a fixed but unknown bottleneck bandwidth called
$\mathit{bw}$ and a controller action $\mathit{send}$ representing the
next sending rate. If the chosen rate is below the bandwidth,  transmission
succeeds. However, if it is at least the bandwidth, a loss occurs. 
The safety objective
has two parts. First, the chosen sending rate must not exceed the true bandwidth
by more than one unit. Second, until the sending rate has reached the bandwidth,
the controller must continue probing upward over time. Thus, the controller must
be aggressive enough to discover the available bandwidth, but not so aggressive
that it overshoots by more than one.
The key difference between the two benchmarks is how loss information is
observed. In the ``no-jitter'' benchmark, loss is observed immediately. In the ``jitter'' benchmark, the loss outcome, stored in a latent
variable $\mathit{currentloss}$, becomes observable only one round later.

\bfpara{Need for different memory abstractions.}
In both benchmarks, a winning strategy must probe the network until it has
learned enough about the hidden bandwidth to choose safe sending rates. Initially,
the controller only has a lower bound on the bandwidth. A successful probe at
rate $r$ shows that the bandwidth is larger than $r$, thereby improving this
lower bound. A loss at rate $r$ shows that the bandwidth is at most $r$, thereby
providing an upper bound.

In the no-jitter benchmark, these lower and upper bounds on \emph{bw} are sufficient for synthesizing a winning strategy.  In particular, if the current
lower bound is $a$, the controller can probe with rate $a+1$. If the probe
succeeds, it learns $\mathit{bw}>a+1$ and raises the lower bound. If the
probe loses, it learns
$
  a < \mathit{bw} \le a+1 
$.
Hence, this simple interval representation suffices.

In contrast, for  the jitter benchmark, maintaining only lower and upper bounds on
$\mathit{bw}$ is not sufficient. Suppose the controller has confirmed a lower bound $m_1$ and then
probes at rate $m_2$. Until the delayed feedback arrives, the controller cannot
classify $m_2$ as either below or above the true bandwidth, so it cannot yet fold
$m_2$ into an interval. Thus, the controller
must remember not only its current bounds, but also the pending probe $m_2$.

\bfpara{Learned memory abstractions and support.}
The memory abstractions learned by \tool{} reflect the distinction above. In the
no-jitter benchmark, the learned epistemic support has an interval interpretation
over the hidden bandwidth. Ignoring  irrelevant variables, a memory value $(a,b)$ represents states satisfying
$ 
  \SafeReach_{\mathsf{nojit}}(a,b,s)
  \equiv
  a < s.\mathit{bw} \le b $
where $b$ may be $+\infty$.

In the jitter benchmark, there is an additional variable that distinguishes different phases of the controller's knowledge. We
write a memory value as $(c,m_1,m_2)$, where $c$ is a phase variable and
$m_1,m_2$ are numeric memory values. The phase $c=1$ denotes the pending-probe
case: $m_1$ is a confirmed lower bound on the bandwidth, while $m_2$ is a probe
whose outcome has not yet become observable. The phase $c=2$ denotes the
bounded-interval case, where $m_1$ and $m_2$ are lower and upper bounds on the
bandwidth.
The learned support relation $  \SafeReach_{\mathsf{jit}}((c,m_1,m_2),s)$  is then given by:
\[
\begin{aligned}
&
  c=1
  \wedge s.\mathit{bw} > m_1
  \wedge
  (s.\mathit{currentloss} \neq 1 \vee s.\mathit{bw} \le m_2)
  \wedge
  (s.\mathit{currentloss} = 1 \vee s.\mathit{bw} > m_2)
\\
&{}\vee
  c=2
  \wedge m_1 < s.\mathit{bw} \le m_2 
\end{aligned}
\]
Thus,  in the pending-probe case, the controller remembers the
probe value $m_2$, and the support relation explains how this probe relates to
the latent loss flag. 

\bfpara{Synthesized policies.}
Using these learned support relations, \tool{} synthesizes policies and memory
updates with a natural interpretation. In the no-jitter benchmark, the synthesized policy probes while only a lower bound is known and
then uses the inferred upper bound:
\[
  \Policy_{\mathsf{nojit}}((a,+\infty)) = a+1
  \qquad
  \Policy_{\mathsf{nojit}}((a,b)) = b 
\]
Writing $\ell$ for the observed loss signal, the memory update is given by:
\[
\MemUpdate_{\mathsf{nojit}}((a,b),\ell) =
\begin{cases}
  (a+1,+\infty)
    & \text{if } b=+\infty \wedge \ell=0 \\
  (a,a+1)
    & \text{if } b=+\infty \wedge \ell=1 \\
  (a,b)
    & \text{if } b\neq +\infty 
\end{cases}
\]

In the jitter benchmark, the  synthesized policy corresponds to:
\[
\Policy_{\mathsf{jit}}(c,m_1,m_2) =
\begin{cases}
  m_2   & \text{if } (c=1 \wedge m_2 > m_1) \lor c = 2 \\
  m_2+1 & \text{if } c=1 \wedge m_2 \le m_1 \\
\end{cases}
\]
When $c=1$ and $m_2$ is ahead of the confirmed lower bound $m_1$, the controller
waits for the pending probe to be resolved rather than issuing a more aggressive
probe. When $m_2 \le m_1$, the pending-probe value does
not exceed the confirmed lower bound, so the controller tries the next rate
$m_2+1$. When $c=2$, the controller has a bounded interval and sends at
the inferred upper bound.

The memory update mirrors the delayed observation model. Let $\ell$ denote the
loss signal that becomes observable in the next observation. The update is
\[
\MemUpdate_{\mathsf{jit}}((c,m_1,m_2),\ell) =
\begin{cases}
  (1,m_2,\Policy_{\mathsf{jit}}(1,m_1,m_2))
    & \text{if } c=1 \wedge \ell=0,\\
  (2,m_1,m_2)
    & \text{if } (c=1 \wedge \ell=1) \lor c = 2
\end{cases}
\]
If the delayed observation reports no loss, then the previous pending probe
$m_2$ is promoted to a confirmed lower bound, and the rate chosen in the current
round becomes the next pending probe. If the delayed observation reports loss,
then $m_2$ becomes an upper bound, and the memory moves to the next
phase. Once the memory is in the second phase, the update preserves
the interval.

\vspace{0.2in}

\idiotbox{RQ4}{
Even closely related games may require  different memory domains, because small changes in observability can change what information a winning strategy must retain. \tool{} can automatically learn  memory abstractions that reflect these differences.
}

\section{Related Work}\label{sec:related}

\bfpara{Perfect-information reactive synthesis.}
Reactive synthesis is classically formulated as a two-player game between a
controller and an adversarial environment. In the perfect-information setting,
the controller observes the full game state before choosing an action. For
safety objectives, memoryless strategies suffice, and the winning region can be
computed by a fixed point, or dually by an attractor computation~\cite{BloemCJ18}. This
game-theoretic view underlies many reactive-synthesis tools, including symbolic
methods for safety, parity, and GR(1)-style specifications~\cite{
PitermanPS06,BloemJPPS12,JobstmannGWB07,PnueliR89}.

For infinite-state systems, the same fixed-point principles have been lifted to
symbolic settings. Prior work develops
abstraction-refinement and constraint-solving techniques for infinite-state
games and control problems with $\omega$-regular or safety objectives
~\cite{HenzingerJM03,BeyeneCPR14}. More recent work improves the scalability of
symbolic infinite-state game solving using acceleration, localized attractor
computation, and modular reasoning~\cite{HeimDimitrovaPOPL24,SchmuckHDN24,
HeimDimitrovaPOPL25,HeimDimitrovaTACAS26}. Another line of work synthesizes
strategies for arithmetic games by generalizing from bounded plays and extracting
symbolic strategies from SMT constraints~\cite{FarzanKincaid18}. Our method for solving the induced memory game builds on all of this prior work on solving perfect-information games.

\bfpara{Finite-state belief-game constructions.}
Under imperfect information, the controller must choose actions based only on its
observation history rather than the actual state. In the finite-state setting,
the classical solution to partial observability is the \emph{knowledge-based
subset construction}, which transforms the game into a perfect-information game
over sets of possible concrete states~\cite{Reif84}. This idea extends to
$\omega$-regular objectives and has led to a rich algorithmic theory of
finite-state games with imperfect information~\cite{ChatterjeeHenzinger07}.
\rev{Kupferman and Vardi~\cite{KupfermanVardi2000} also study
synthesis with incomplete information from temporal-logic
specifications using automata-theoretic techniques.}
Related
finite-state synthesis work studies epistemic-temporal specifications, where
desired behavior may refer directly to what agents know along an execution
~\cite{MeydenW05}. However, belief-style constructions are expensive even for
finite-state games, where the number of information sets may be exponential in
the state space. Antichain and lattice-based representations mitigate this
blow-up by exploiting partial orders over information sets~\cite{DeWulfDR06},
but they still fundamentally reason over sets of possible states. Our approach
avoids constructing the full belief game and instead learns a symbolic memory
abstraction and an epistemic support relation that compactly describe
safety-relevant uncertainty.

\bfpara{Abstraction refinement for infinite-state imperfect-information games.}
The closest line of work is the abstraction-refinement approach of Dimitrova and
Finkbeiner for infinite-state games with incomplete information~\cite{
DimitrovaFinkbeinerFSTTCS08,DimitrovaThesis}. Their method combines predicate
abstraction with the knowledge-based subset construction to obtain a \emph{finite}
perfect-information abstraction of an infinite-state incomplete-information game.
If the environment wins the abstract game, a counterexample tree is analyzed and
the abstraction is refined. This is closely related to our
goal of avoiding explicit construction of the concrete belief game. However, their
CEGAR construction assumes that the controller has a finite action alphabet. By
contrast, our induced memory game is represented symbolically, where memory values may
range over infinite domains, and controller actions are synthesized from logical
constraints.

Dimitrova and Finkbeiner later extend this abstraction-refinement perspective
to timed control with partial observation~\cite{DimitrovaFinkbeinerFORMATS12,
DimitrovaThesis}. In that setting, the controller must decide not only which
 action to take, but also \emph{when} to take it. In particular, because time is modeled as a
dense quantity, there are infinitely many possible waiting times. Their approach
handles this source of infiniteness by synthesizing a finite set of
\emph{action points}, which are timing choices that are sufficient for the controller
to consider. This reduces the real-time control problem to one with finitely many
relevant timing choices. Our setting is different in that the infinite action space does
not come \emph{only} from choosing when to act, but from general real-valued control
commands. Rather than
discovering a finite set of representative actions, our algorithm instead reasons
symbolically about the existence of safe actions in the background theory.

\bfpara{Observation design for partial-information synthesis.}
A related line of work treats the controller's observation interface as part of
the synthesis problem. For example, Fu et al. study how sensing limitations
affect realizability in partial-information reactive  synthesis and use
abstraction refinement to identify sensing modalities that should be added
~\cite{FuDimitrovaTopcuACC14}. Related work in active sensing incorporates
sensing actions into reactive synthesis so that the controller can reduce
uncertainty online~\cite{FuTopcuActiveSensing14}. Another line of work
studies sensor selection and observation masks for   systems under
partial observation~\cite{JiangKumarGarciaTAC03}. These techniques study what the
system should observe, either by changing the sensor interface or by taking
additional sensing actions. Our setting is complementary in that the observation
interface is fixed, and we must learn what information from the observation
history should be retained as internal memory.

\bfpara{Estimator-based synthesis.}
Estimator-based reactive synthesis decomposes partial-information synthesis into
two steps: first, it constructs a state-estimation component that summarizes
relevant information about unobserved variables from the observation history;
second, it performs complete-information synthesis for a controller that uses
this estimate~\cite{EhlersT15}. This is complementary to our approach. Rather
than constructing a separate estimator interface, \tool{} learns a symbolic
memory abstraction and its epistemic support relation.

\bfpara{Belief-state programming.}
Belief programming provides a programming model and verification logic for
controllers that explicitly maintain and query belief states at runtime
~\cite{AtkinsonCarbin2020PartialObservability}. Subsequent work extends this
idea to probabilistic partially observable environments, where the belief state
is a probability distribution over possible environment states~\cite{
GuertlerKaminski2025PBLIMP}. These approaches are complementary to ours in that they
provide languages and logics for writing and verifying belief-manipulating
programs. In contrast, \tool{} synthesizes the controller's memory representation,
memory update, and control policy from a symbolic partially observable safety
game.

\bfpara{Domain-specific synthesis under partial observability.}
Our congestion-control benchmarks are related to recent work on provably
performant congestion control~\cite{agarwal_nsdi24_belief_cc}. That work gives a
domain-specific synthesis framework for congestion-control algorithms using
CCA templates, network-specific belief computations, and performance properties
tailored to congestion control. In contrast, \tool{} takes a general symbolic
partially observable safety game as input and is not specialized to a particular
network model or controller template.

\rev{\bfpara{Distributed synthesis.}
Partial information also plays a central role in
distributed synthesis, where individual processes
must make decisions without observing the complete
system state. Kupferman and Vardi~\cite{KupfermanVardiDistributed}
study the synthesis of distributed systems from
temporal specifications. More recently, Finkbeiner
et al.~\cite{FinkbeinerMetzgerMoses2022} use
information-flow assumptions to guide compositional
synthesis. While these approaches focus on
synthesizing distributed implementations under
architectural information constraints, our work
focuses on automatically inferring symbolic memory
abstractions for infinite-state safety games.}

\bfpara{POMDPs.}
Partially observable Markov decision processes (POMDPs) are a classical model
for sequential decision making under uncertainty~\cite{KaelblingLC98}. POMDP
solvers typically reason over belief distributions and synthesize policies that
optimize expected reward or the probability of satisfying a specification
~\cite{PineauGHT03,SpaanVlassis05,ChatterjeeChmelikTracol13}. \rev{Another line of work studies qualitative objectives,
including almost-sure satisfaction of safety and other
$\omega$-regular properties~\cite{ChatterjeeDoyenHenzinger2010}.
Junges et al.~\cite{Junges2018} and subsequent work
study the synthesis of finite-state controllers
for POMDPs.}
Our setting is
different in that  we consider adversarial safety games over symbolic 
infinite-state systems.

\section{Conclusion and Future Work}
\label{sec:conclusion}

This paper presented a new synthesis algorithm for partially observable safety
games over symbolic, infinite-state models with real-valued action spaces. The
main idea is to algorithmically separate the problem of determining \emph{what the controller
should remember} from the problem of determining \emph{how the controller should
act}. To this end, our algorithm first learns a symbolic memory abstraction
together with an epistemic support relation that gives semantic meaning to
memory values in terms of the concrete states they may represent. It then uses
this support relation to construct an induced perfect-information safety game
over memories, from which the control policy and memory-update function can be
synthesized using symbolic fixpoint reasoning.

We implemented this approach in a prototype called \tool{} and evaluated it on benchmarks from UAV
control, robot navigation, lander descent, and congestion control. Our
experimental results show that symbolic memory abstractions can make partially
observable synthesis practical on challenging benchmarks with infinite state
spaces and real-valued controller actions. Our ablation study further shows that
the key algorithmic ingredients of our method are important for solving these benchmarks in practice.

There are several directions for future work. One important direction is to
improve scalability. Our current refinement strategy for forward exploration is
deliberately simple, and more targeted refinements could help learn useful
support relations with fewer iterations.    Another promising direction is to extend this method beyond
safety, for example to reachability, liveness, or quantitative properties, where
the memory abstraction must support not only safe action choices but also
progress or optimization objectives.

\section{Data-Availability Statement}

Our artifact, including the \tool{} implementation, benchmark suite,
and evaluation scripts, is available on Zenodo~\cite{pan_2026_24568948}.

\section*{Acknowledgments}
We thank the anonymous reviewers for their thoughtful and constructive feedback.
This work was conducted in a research group supported by NSF awards CCF-1918889, CNS-1908304, CCF-1901376, CNS-2120696, CCF-2210831, CCF-2319471, CCF-2422130, and CCF-2403211, as well as DARPA awards under agreements HR00112590133 and HR0011262E020, and a Renaissance Philanthropy award.

\bibliographystyle{ACM-Reference-Format}
\bibliography{main}

@inproceedings{PnueliRosner,
  author    = {Amir Pnueli and Roni Rosner},
  title     = {On the Synthesis of a Reactive Module},
  booktitle = {Proceedings of the 16th {ACM} {SIGPLAN}{-}{SIGACT} Symposium on
               Principles of Programming Languages ({POPL})},
  pages     = {179--190},
  publisher = {{ACM}},
  year      = {1989}
}

@inproceedings{EhlersGR1,
  author    = {R{\"u}diger Ehlers},
  title     = {Unbeast: Symbolic Bounded Synthesis},
  booktitle = {Tools and Algorithms for the Construction and Analysis of Systems ({TACAS})},
  series    = {Lecture Notes in Computer Science},
  volume    = {6605},
  pages     = {272--275},
  publisher = {Springer},
  year      = {2011}
}

@inproceedings{DimitrovaFinkbeinerFSTTCS08,
  author    = {Rayna Dimitrova and Bernd Finkbeiner},
  title     = {Abstraction Refinement for Games with Incomplete Information},
  booktitle = {Foundations of Software Technology and Theoretical Computer Science (FSTTCS 2008)},
  series    = {Leibniz International Proceedings in Informatics (LIPIcs)},
  volume    = {2},
  pages     = {175--186},
  publisher = {Schloss Dagstuhl--Leibniz-Zentrum f{\"u}r Informatik},
  year      = {2008}
}

@article{reif84,
  author  = {Reif, John H.},
  title   = {The Complexity of Two-Player Games of Incomplete Information},
  journal = {Journal of Computer and System Sciences},
  volume  = {29},
  number  = {2},
  pages   = {274--301},
  year    = {1984},
  doi     = {10.1016/0022-0000(84)90034-5}
}

@article{ChatterjeeHenzinger07,
  author  = { Krishnendu Chatterjee and Laurent Doyen and Thomas A. Henzinger and Jean-Fran{\c{c}}ois Raskin},
  title   = {Algorithms for Omega-Regular Games with Imperfect Information},
  journal = {Logical Methods in Computer Science},
  volume  = {3},
  number  = {3},
  year    = {2007},
  doi     = {10.2168/LMCS-3(3:4)2007},
  url     = {https://doi.org/10.2168/LMCS-3(3:4)2007}
}

@inproceedings{DeWulfDR06,
  author    = {Martin De Wulf and Laurent Doyen and Jean-Fran{\c{c}}ois Raskin},
  title     = {A Lattice Theory for Solving Games of Imperfect Information},
  booktitle = {Hybrid Systems: Computation and Control (HSCC 2006)},
  series    = {Lecture Notes in Computer Science},
  volume    = {3927},
  pages     = {153--168},
  publisher = {Springer},
  year      = {2006},
  doi       = {10.1007/11730637_13},
  url       = {https://doi.org/10.1007/11730637_13}
}

@inproceedings{HenzingerJM03,
  author    = {Thomas A. Henzinger and Ranjit Jhala and Rupak Majumdar},
  title     = {Counterexample-Guided Control},
  booktitle = {Automata, Languages and Programming (ICALP 2003)},
  series    = {Lecture Notes in Computer Science},
  volume    = {2719},
  pages     = {886--902},
  publisher = {Springer},
  year      = {2003},
  doi       = {10.1007/3-540-45061-0_69},
  url       = {https://doi.org/10.1007/3-540-45061-0_69}
}

@inproceedings{EhlersT15,
  author    = {R{\"u}diger Ehlers and Ufuk Topcu},
  title     = {Estimator-based Reactive Synthesis under Incomplete Information},
  booktitle = {Proceedings of the 18th International Conference on Hybrid Systems: Computation and Control (HSCC 2015)},
  year      = {2015},
  url       = {https://dblp.org/rec/conf/hybrid/EhlersT15}
}

@inproceedings{MeydenW05,
  author    = {Ronen van der Meyden and Thomas Wilke},
  title     = {Synthesis of Distributed Systems from Knowledge-Based Specifications},
  booktitle = {CONCUR 2005 -- Concurrency Theory},
  series    = {Lecture Notes in Computer Science},
  volume    = {3653},
  pages     = {332--347},
  publisher = {Springer},
  year      = {2005},
  doi       = {10.1007/11539452_26},
  url       = {https://doi.org/10.1007/11539452_26}
}

@article{AtkinsonCarbin2020PartialObservability,
  author  = {Eric Atkinson and Michael Carbin},
  title   = {Programming and Reasoning with Partial Observability},
  journal = {Proceedings of the ACM on Programming Languages},
  volume  = {4},
  number  = {OOPSLA},
  articleno = {200},
  year    = {2020},
  month   = nov,
  doi     = {10.1145/3428268},
  url     = {https://doi.org/10.1145/3428268}
}

@inproceedings{PitermanPS06,
  author    = {Nir Piterman and Amir Pnueli and Yaniv Sa'ar},
  title     = {Synthesis of Reactive(1) Designs},
  booktitle = {Verification, Model Checking, and Abstract Interpretation (VMCAI 2006)},
  series    = {Lecture Notes in Computer Science},
  volume    = {3855},
  pages     = {364--380},
  publisher = {Springer},
  year      = {2006},
  doi       = {10.1007/11609773_24}
}

@inproceedings{JobstmannGWB07,
  author    = {Barbara Jobstmann and Stefan J. Galler and Martin Weiglhofer and Roderick Bloem},
  title     = {Anzu: A Tool for Property Synthesis},
  booktitle = {Computer Aided Verification (CAV)},
  series    = {Lecture Notes in Computer Science},
  volume    = {4590},
  publisher = {Springer},
  year      = {2007},
  pages     = {258--262}
}

@inproceedings{PnueliR89,
  author    = {Amir Pnueli and Roni Rosner},
  title     = {On the Synthesis of an Asynchronous Reactive Module},
  booktitle = {Proceedings of the 16th International Colloquium on Automata, Languages and Programming (ICALP)},
  series    = {Lecture Notes in Computer Science},
  publisher = {Springer},
  year      = {1989},
  pages     = {652--671}
}

@article{FarzanKincaid18,
  author  = {Azadeh Farzan and Zachary Kincaid},
  title   = {Strategy Synthesis for Linear Arithmetic Games},
  journal = {Proceedings of the ACM on Programming Languages},
  volume  = {2},
  number  = {POPL},
  year    = {2018},
  articleno = {61},
  numpages  = {30},
  doi     = {10.1145/3158149}
}

@inproceedings{BeyeneCPR14,
  author    = {Tewodros A. Beyene and Swarat Chaudhuri and Corneliu Popeea and Andrey Rybalchenko},
  title     = {A Constraint-Based Approach to Solving Games on Infinite Graphs},
  booktitle = {Proceedings of the 41st ACM SIGPLAN-SIGACT Symposium on Principles of Programming Languages (POPL)},
  publisher = {ACM},
  year      = {2014},
  pages     = {221--234}
}

@article{BloemJPPS12,
  author  = {Roderick Bloem and Barbara Jobstmann and Nir Piterman and Amir Pnueli and Yaniv Sa'ar},
  title   = {Synthesis of Reactive(1) Designs},
  journal = {Journal of Computer and System Sciences},
  volume  = {78},
  number  = {3},
  pages   = {911--938},
  year    = {2012},
  doi     = {10.1016/j.jcss.2011.08.007}
}

@incollection{BloemCJ18,
  author    = {Roderick Bloem and Krishnendu Chatterjee and Barbara Jobstmann},
  title     = {Graph Games and Reactive Synthesis},
  booktitle = {Handbook of Model Checking},
  editor    = {Edmund M. Clarke and Thomas A. Henzinger and Helmut Veith and Roderick Bloem},
  publisher = {Springer},
  year      = {2018},
  doi       = {10.1007/978-3-319-10575-8_27}
}

@inproceedings{agarwal_nsdi24_belief_cc,
  author    = {Agarwal, Anup and Arun, Venkat and Ray, Devdeep and Martins, Ruben and Seshan, Srinivasan},
  title     = {Towards Provably Performant Congestion Control},
  booktitle = {21st USENIX Symposium on Networked Systems Design and Implementation (NSDI 24)},
  year      = {2024},
  month     = apr,
  pages     = {951--978},
  publisher = {USENIX Association},
  address   = {Santa Clara, CA},
  url       = {https://www.usenix.org/conference/nsdi24/presentation/agarwal-anup}
}

@inproceedings{DilligSAS2010Simplify,
  author       = {Isil Dillig and Thomas Dillig and Alex Aiken},
  title        = {Small Formulas for Large Programs: On-Line Constraint Simplification in Scalable Static Analysis},
  booktitle    = {Static Analysis - 17th International Symposium (SAS 2010), Perpignan, France, September 14--16, 2010. Proceedings},
  editor       = {Radhia Cousot and Matthieu Martel},
  series       = {Lecture Notes in Computer Science},
  volume       = {6337},
  pages        = {236--252},
  publisher    = {Springer},
  year         = {2010},
  doi          = {10.1007/978-3-642-15769-1_15}
}

@MISC{smtlib,
  author =	 {Clark Barrett and Pascal Fontaine and Cesare Tinelli},
  title =	 {{The Satisfiability Modulo Theories Library (SMT-LIB)}},
  howpublished = {{\tt www.SMT-LIB.org}},
  year =	 2016,
}

@inproceedings{z3,
author = {de Moura, Leonardo and Bjørner, Nikolaj},
title = {Z3: an efficient SMT solver},
booktitle = {2008 Tools and Algorithms for Construction and Analysis of Systems},
year = {2008},
month = {March},
publisher = {Springer, Berlin, Heidelberg},
url = {https://www.microsoft.com/en-us/research/publication/z3-an-efficient-smt-solver/},
pages = {337-340},
}

@inproceedings{DimitrovaFinkbeinerFORMATS12,
  author    = {Rayna Dimitrova and Bernd Finkbeiner},
  title     = {Counterexample-Guided Synthesis of Observation Predicates},
  booktitle = {Formal Modeling and Analysis of Timed Systems},
  editor    = {Marcin Jurdzi{\'n}ski and Dejan Ni{\v{c}}kovi{\'c}},
  series    = {Lecture Notes in Computer Science},
  volume    = {7595},
  pages     = {107--122},
  publisher = {Springer},
  year      = {2012},
  doi       = {10.1007/978-3-642-33365-1_9}
}

@inproceedings{deAlfaroHM01,
  author    = {Luca de Alfaro and Thomas A. Henzinger and Rupak Majumdar},
  title     = {Symbolic Algorithms for Infinite-State Games},
  booktitle = {CONCUR 2001},
  series    = {LNCS},
  volume    = {2154},
  pages     = {536--550},
  publisher = {Springer},
  year      = {2001}
}

@article{HeimDimitrovaPOPL24,
  author  = {Philippe Heim and Rayna Dimitrova},
  title   = {Solving Infinite-State Games via Acceleration},
  journal = {Proceedings of the ACM on Programming Languages},
  volume  = {8},
  number  = {POPL},
  pages   = {1696--1726},
  year    = {2024},
  doi     = {10.1145/3632899}
}

@inproceedings{SchmuckHDN24,
  author    = {Anne-Kathrin Schmuck and Philippe Heim and Rayna Dimitrova and Satya Prakash Nayak},
  title     = {Localized Attractor Computations for Infinite-State Games},
  booktitle = {Computer Aided Verification - 36th International Conference, CAV 2024},
  series    = {Lecture Notes in Computer Science},
  volume    = {14683},
  pages     = {135--158},
  publisher = {Springer},
  year      = {2024},
  doi       = {10.1007/978-3-031-65633-0_7}
}

@article{HeimDimitrovaPOPL25,
  author  = {Philippe Heim and Rayna Dimitrova},
  title   = {Translation of Temporal Logic for Efficient Infinite-State Reactive Synthesis},
  journal = {Proceedings of the ACM on Programming Languages},
  volume  = {9},
  number  = {POPL},
  pages   = {1536--1567},
  year    = {2025},
  doi     = {10.1145/3704888}
}

@inproceedings{HeimDimitrovaTACAS26,
  author    = {Philippe Heim and Rayna Dimitrova},
  title     = {Modular Attractor Acceleration in Infinite-State Games},
  booktitle = {Tools and Algorithms for the Construction and Analysis of Systems - 32nd International Conference, TACAS 2026},
  series    = {Lecture Notes in Computer Science},
  pages     = {398--418},
  publisher = {Springer},
  year      = {2026},
  doi       = {10.1007/978-3-032-22752-2_21}
}

@phdthesis{DimitrovaThesis,
  author = {Rayna Dimitrova},
  title  = {Synthesis and Control of Infinite-State Systems with Partial Observability},
  school = {Saarland University},
  year   = {2013},
  type   = {Ph.D. thesis}
}

@inproceedings{FuDimitrovaTopcuACC14,
  author    = {Jie Fu and Rayna Dimitrova and Ufuk Topcu},
  title     = {Abstractions and Sensor Design in Partial-Information, Reactive Controller Synthesis},
  booktitle = {Proceedings of the 2014 American Control Conference},
  pages     = {2297--2304},
  publisher = {IEEE},
  year      = {2014},
  doi       = {10.1109/ACC.2014.6858842}
}

@article{FuTopcuActiveSensing14,
  author  = {Jie Fu and Ufuk Topcu},
  title   = {Integrating Active Sensing into Reactive Synthesis with Temporal Logic Constraints under Partial Observations},
  journal = {CoRR},
  volume  = {abs/1410.0083},
  year    = {2014},
  eprint  = {1410.0083},
  archivePrefix = {arXiv}
}

@article{JiangKumarGarciaTAC03,
  author  = {Shengbing Jiang and Ratnesh Kumar and H. E. Garcia},
  title   = {Optimal Sensor Selection for Discrete-Event Systems with Partial Observation},
  journal = {IEEE Transactions on Automatic Control},
  volume  = {48},
  number  = {3},
  pages   = {369--381},
  year    = {2003},
  doi     = {10.1109/TAC.2003.809144}
}

@article{GuertlerKaminski2025PBLIMP,
  author        = {Tobias G{\"u}rtler and Benjamin Lucien Kaminski},
  title         = {Programming and Reasoning in Partially Observable Probabilistic Environments},
  journal       = {CoRR},
  volume        = {abs/2506.13491},
  year          = {2025},
  eprint        = {2506.13491},
  archivePrefix = {arXiv}
}

@article{KaelblingLC98,
  author  = {Leslie Pack Kaelbling and Michael L. Littman and Anthony R. Cassandra},
  title   = {Planning and Acting in Partially Observable Stochastic Domains},
  journal = {Artificial Intelligence},
  volume  = {101},
  number  = {1--2},
  pages   = {99--134},
  year    = {1998},
  doi     = {10.1016/S0004-3702(98)00023-X}
}

@inproceedings{PineauGHT03,
  author    = {Joelle Pineau and Geoff Gordon and Sebastian Thrun},
  title     = {Point-Based Value Iteration: An Anytime Algorithm for {POMDP}s},
  booktitle = {Proceedings of the Eighteenth International Joint Conference on Artificial Intelligence},
  pages     = {1025--1030},
  publisher = {Morgan Kaufmann},
  year      = {2003}
}

@article{SpaanVlassis05,
  author  = {Matthijs T. J. Spaan and Nikos Vlassis},
  title   = {Perseus: Randomized Point-Based Value Iteration for {POMDP}s},
  journal = {Journal of Artificial Intelligence Research},
  volume  = {24},
  pages   = {195--220},
  year    = {2005},
  doi     = {10.1613/jair.1659}
}

@article{ChatterjeeChmelikTracol13,
  author  = {Krishnendu Chatterjee and Martin Chmelik and Mathieu Tracol},
  title   = {What Is Decidable about Partially Observable Markov Decision Processes with {$\omega$}-Regular Objectives},
  journal = {Journal of Computer and System Sciences},
  volume  = {82},
  number  = {5},
  pages   = {878--911},
  year    = {2016},
  doi     = {10.1016/j.jcss.2016.02.009}
}

@incollection{KupfermanVardi2000,
  author    = {Orna Kupferman and Moshe Y. Vardi},
  title     = {Synthesis with Incomplete Information},
  booktitle = {Advances in Temporal Logic},
  editor    = {Howard Barringer and Michael Fisher and
               Dov M. Gabbay and Graham Gough},
  series    = {Applied Logic Series},
  volume    = {16},
  pages     = {109--127},
  year      = {2000},
  publisher = {Springer},
  doi       = {10.1007/978-94-015-9586-5_6}
}

@inproceedings{KupfermanVardiDistributed,
  author    = {Orna Kupferman and Moshe Y. Vardi},
  title     = {Synthesizing Distributed Systems},
  booktitle = {Proceedings of the 16th Annual IEEE
               Symposium on Logic in Computer Science
               (LICS)},
  pages     = {389--398},
  year      = {2001},
  publisher = {IEEE Computer Society}
}

@inproceedings{FinkbeinerMetzgerMoses2022,
  author    = {Bernd Finkbeiner and Niklas Metzger and
               Yoram Moses},
  title     = {Information Flow Guided Synthesis},
  booktitle = {Computer Aided Verification (CAV)},
  editor    = {Sharon Shoham and Yakir Vizel},
  series    = {Lecture Notes in Computer Science},
  volume    = {13372},
  pages     = {505--525},
  year      = {2022},
  publisher = {Springer},
  doi       = {10.1007/978-3-031-13188-2_25}
}

@inproceedings{ChatterjeeDoyenHenzinger2010,
  author    = {Krishnendu Chatterjee and Laurent Doyen and
               Thomas A. Henzinger},
  title     = {Qualitative Analysis of Partially-Observable
               Markov Decision Processes},
  booktitle = {Mathematical Foundations of Computer Science
               (MFCS)},
  editor    = {Petr Hlin{\v{e}}n{\'y} and Anton{\'i}n Ku{\v{c}}era},
  series    = {Lecture Notes in Computer Science},
  volume    = {6281},
  pages     = {258--269},
  year      = {2010},
  publisher = {Springer},
  doi       = {10.1007/978-3-642-15155-2_24}
}

@inproceedings{Junges2018,
  author    = {Sebastian Junges and Nils Jansen and
               Ralf Wimmer and Tim Quatmann and
               Leonore Winterer and Joost-Pieter Katoen and
               Bernd Becker},
  title     = {Finite-State Controllers of {POMDPs}
               using Parameter Synthesis},
  booktitle = {Proceedings of the 34th Conference on
               Uncertainty in Artificial Intelligence (UAI)},
  pages     = {519--529},
  year      = {2018},
  publisher = {AUAI Press},
  url       = {https://www.auai.org/uai2018/proceedings/papers/195.pdf}
}

@misc{extended_paper,
      title={Solving Partially Observable Safety Games via Symbolic Memory Abstractions (TBD: update this entry)},
      author={Jia Pan},
      year={2026},
      eprint={2508.15109},
      archivePrefix={arXiv},
      primaryClass={cs.PL},
      url={https://arxiv.org/abs/2508.15109}, 
}

@software{pan_2026_24568948,
  author       = {Pan, Jia and
                  Wang, Ziteng and
                  Arun, Venkat and
                  McMillan, Kenneth and
                  Dillig, Isil},
  title        = {Verena: Reproduction Package for "Solving
                   Partially Observable Safety Games via Symbolic
                   Memory Abstractions"
                  },
  month        = nov,
  year         = 2026,
  publisher    = {Zenodo},
  doi          = {10.5281/zenodo.24568947},
  url          = {https://doi.org/10.5281/zenodo.24568947},
}

\ifextended
\newpage
\appendix
\section{Why the Motivating Example Does Not Admit Memoryless Strategy}
\label{app:motivating}

We show by contradiction that this example does not admit a memoryless winning strategy.

At $x=0$, the controller has not observed any altitude information. Hence its
first vertical adjustment is some fixed value $a_0\in[-1,1]$. After this move,
the robot is at $x=1$, and its altitude lies in the interval $[a_0,a_0+6]$.

At $x=1$, the robot observes the beacon reading $z=y+b$. Since the action at
$x=1$ may depend on this observation, write it as $a_1(z)\in[-1,1]$. After the
robot leaves the beacon, however, the beacon reading is no longer observable.
Thus, at positions $x=2,\ldots,6$, a memoryless strategy can choose actions only
as a function of the current position. These actions are therefore fixed values,
independent of $z$. Let
$ 
  C = a_2+\cdots+a_6
$
denote their sum.

Now consider any beacon reading
$ 
  z \in [a_0+1,a_0+5].
$
For every such $z$, there are two possible hidden states at $x=1$ that produce
the same observation $z$. Namely, if $b=+1$, then $y=z-1$, and if $b=-1$, then
$y=z+1$. Both altitudes lie in $[a_0,a_0+6]$ by the choice of $z$, so both
states are reachable and indistinguishable to the controller.

Since the strategy is winning, it must safely handle both hidden states. At the
wall, the altitude is
\[
  y_{\mathit{wall}} = z-b+a_1(z)+C.
\]
If $b=+1$, safety requires the robot to pass through the lower door:
\[
  0 < z-1+a_1(z)+C < 1.
\]
If $b=-1$, safety requires the robot to pass through the upper door:
\[
  2 < z+1+a_1(z)+C < 3.
\]
Both inequalities are equivalent to
$ 
  1 < z+a_1(z)+C < 2.
$
Equivalently, writing $r=z+C$, the action at the beacon must choose
$a_1(z)\in[-1,1]$ so that
$ 
  r+a_1(z) \in (1,2).
$
This is possible only when $r\in(0,3)$. Thus, for every
$z\in[a_0+1,a_0+5]$, we must have
$ 
  z+C \in (0,3).
$
But as $z$ ranges over $[a_0+1,a_0+5]$, the value $z+C$ ranges over the interval
$ 
  [a_0+1+C,\; a_0+5+C],
$
which has length $4$. But no interval of length $4$ can be contained in the interval
$(0,3)$, yielding a contradiction.

\section{ Strategy Extraction}
\label{appendix:strategy-extraction}

\begin{algorithm}[t]
  \caption{\textsc{ExtractStrategy}}
  \label{alg:extract-strategy}
  \begin{algorithmic}[1]
    \State $\Skel \gets \textsc{ExtractSkeleton}(P)$
    \ForAll{$(\Act_i,\NextMem_i) \in \Skel$}
      \State $\gamma_{\NextMem_i}(m,o)
        \gets \textsc{QE}_{s,s'}\bigl(
              P(m,o,s,s',\Act_i(m),\NextMem_i(m,o))\bigr)$
    \EndFor
    \ForAll{distinct $\Act$ with $(\Act,\NextMem) \in \Skel$}
      \State $\Updates(\Act) \gets
        \{\, \NextMem \mid (\Act,\NextMem) \in \Skel \,\}$
      \State $\gamma_{\Act}(m)
        \gets \textsc{QE}_{o,s,s'}\Bigl(
              \bigvee_{\NextMem \in \Updates(\Act)}
                P(m,o,s,s',\Act(m),\NextMem(m,o))\Bigr)$
    \EndFor
    \State $\Policy(m) \gets \mathbf{choose}\ \Act(m)
            \text{ such that } \gamma_{\Act}(m)$
    \State $\MemUpdate(m,o) \gets \mathbf{choose}\ \NextMem(m,o)
            \text{ such that } \gamma_{\NextMem}(m,o)$
    \State \Return $\langle \Mem,\MemInit,\MemUpdate,\Policy \rangle$

  \end{algorithmic}
\end{algorithm}

In our implementation, we solve for the unknown functions $(\Policy,\MemUpdate)$
using a direct adaptation of the
strategy-extraction technique of {Farzan et al.}~\cite{FarzanKincaid18}
for linear arithmetic (Algorithm~\ref{alg:extract-strategy}).
This algorithm  takes as input a formula with quantifiers and constructs a
finite family of \emph{templates} for the existential choices:
in our setting, these are affine terms of the form
  $\Act_i(m)$ and $\NextMem_i(m,o)$
that propose how to pick the action and updated memory.  Intuitively, each
template $(\Act_i,\NextMem_i)$ describes one ``mode'' of the controller's
behavior.

Then, to turn these templates into concrete implementations of $\Policy$ and
$\MemUpdate$, we perform   post-processing  based on quantifier
elimination.
Given the skeleton $\Skel$ returned by the extractor, we first compute, for each pair,
a \emph{local} guard:
\[
  \gamma_{\NextMem_i}(m,o)
     \equiv
    \textsc{QE}_{s,s'}\bigl(
      P(m,o,s,s',\Act_i(m),\NextMem_i(m,o))
    \bigr)
\]

Next, for each distinct action template $\Act$, we collect its associated
update templates
$\Updates(\Act) = \{\,\NextMem \mid (\Act,\NextMem) \in \Skel\,\}$
and compute an \emph{action} guard:
\[
  \gamma_{\Act}(m)
   \equiv 
    \textsc{QE}_{o,s,s'}\Bigl(
      \bigvee_{\NextMem \in \Updates(\Act)}
        P(m,o,s,s',\Act(m),\NextMem(m,o))
    \Bigr),
\]
which holds for memories $m$ from which $\Act$ admits at least one
safe update in $\Updates(\Act)$.
We then define $\Policy$ and $\MemUpdate$ by guarded clauses:
for any $m$ with $\gamma_{\Act}(m)$, we  set $\Policy(m) = \Act(m)$;
for any $(m,o)$ with $\gamma_{\NextMem}(m,o)$, we  set
$\MemUpdate(m,o) = \NextMem(m,o)$.
Thus the guards $\gamma_{\Act}$ and $\gamma_{\NextMem}$ carve the state
space into regions on which these templates are safe.

\subsection{Incremental Belief Exploration} \label{appendix:incremental-belief-exploration} Algorithm~\ref{alg:incremental-explore-beliefs} gives the incremental version of \textsc{ExploreBeliefs} used by \tool{}. The key difference from Algorithm~\ref{alg:explore-beliefs} is that it maintains an additional predicate $\mathsf{Live}_k(\ohist)$, which records those histories that are still worth exploring. Concretely, we say that a history is \emph{live} if it has not been ruled out by the winning condition check at a smaller horizon.

\begin{algorithm}[t]
  \caption{\textsc{IncrementalExploreBeliefs}$(\Init,\Model,\Safe,h)$}
  \label{alg:incremental-explore-beliefs}
  \begin{algorithmic}[1]
    \State $B_0(\epsilon,s) \gets \Init(s)\wedge\Safe(s)$
    \State $\mathsf{Live}_0(\epsilon) \gets \exists s.\,B_0(\epsilon,s)$

    \For{$r = 1$ \textbf{to} $h$}
      \State $B_r(\ohist\cdot a\cdot o,\,s') \gets
        \Safe(s')\wedge
        \mathsf{Live}_{r-1}(\ohist)\wedge
        \exists s.\, B_{r-1}(\ohist,s)\wedge\Model(s,a,s',o)$
      \State $\mathsf{Live}_r(\ohist) \gets \exists s.\,B_r(\ohist,s)$

      \State $\mathsf{Win}_r(\ohist) \gets \mathsf{Live}_r(\ohist)$
      \For{$k = r-1$ \textbf{downto} $0$}
        \State $\mathsf{Win}_k(\ohist) \gets
          \mathsf{Live}_k(\ohist)\wedge
          \exists a.\forall s,s',o.
          \bigl(B_k(\ohist,s)\wedge\Model(s,a,s',o)\bigr)
          \Rightarrow{}
          \bigl(\Safe(s')\wedge
          \mathsf{Win}_{k+1}(\ohist\cdot a\cdot o)\bigr)$
      \EndFor

      \If{$\neg\,\mathsf{Win}_0(\epsilon)$}
        \State \Return \textsc{Failure}
      \EndIf

      \For{$k = 0$ \textbf{to} $r$}
        \State $\mathsf{Live}_k(\ohist) \gets
          \mathsf{Live}_k(\ohist)\wedge\mathsf{Win}_k(\ohist)$
      \EndFor
    \EndFor

    \State \Return
      $[\,B_0\wedge\mathsf{Live}_0,\ldots,
         B_h\wedge\mathsf{Live}_h\,]$
  \end{algorithmic}
\end{algorithm}

The algorithm proceeds by increasing the horizon one step at a time. At
iteration $r$, it first computes the beliefs at depth $r$ from the beliefs at
depth $r-1$ (line~4). The conjunct $\mathsf{Live}_{r-1}(\ohist)$ is the key
optimization: histories that have already been ruled out at an earlier horizon
are not expanded further. 
After extending the beliefs, the algorithm computes the winning predicates for
the current horizon $r$. The base case is
$\mathsf{Win}_r(\ohist)\equiv \mathsf{Live}_r(\ohist)$, rather than simply
$\top$, because we only want to consider live histories. For
earlier depths, $\mathsf{Win}_k(\ohist)$ is also restricted to live
histories.

If the empty history is not winning, then there is no strategy that can guarantee
safety even for the bounded horizon $r$, so the algorithm returns
\textsc{Failure}. Otherwise, the algorithm updates the live predicate by
conjoining it with the newly computed winning predicate. Hence,
any history that is safe so far but cannot be extended safely to the current
horizon is removed from future exploration. After the final iteration, the
algorithm returns the beliefs restricted to histories that remain live through
horizon $h$.

\subsection{Symbolic representation of beliefs}

To represent beliefs $B(\ohist,s)$ symbolically, we require a symbolic representation of observation histories. This problem is somewhat similar to encoding arrays of integers, if you wished to eliminate the theory of arrays. 
Although histories are in principle sequences $a_0,o_1,a_1,o_2,\ldots,a_{k-1},o_{k}$ for arbitrary $k$, we have to represent beliefs only for histories of length $k \leq h$, where $h$ is the exploration horizon. For this reason, we can represent a history $\omega$ by a fixed tuple of terms $(h, c, a_0,o_1,a_1,o_2,\ldots,a_{h-1},o_{h})$ where $h$ is a numeric constant giving the maximum length of the term $c$ is a term bounded in the range $[0,h]$ representing the actual length of $\omega$. This allows us to represent any history up to length $h$. To use this representation, a history-valued variable must have a fixed upper length bound. For example, one cannot write $\exists \pi.\ \omega = \pi \cdot a \cdot o$ but it is possible to write $\exists \pi \in \mathcal{H}_h.\ \omega = \pi \cdot a \cdot o$. In this case, variable $\pi$ can be encoded as a tuple with a fixed value of $h$, in which $c,a_i,o_i$ are fresh variables of the appropriate types. 

With this representation, we can encode $\pi\cdot a\cdot o$ as:

\begin{equation*}
\begin{split}
(\pi.h+1,\pi.c+1, & \mathit{ite}(\pi.c=0,a,\pi.a_0),\mathit{ite}(\pi.c=0,o,\pi.o_1),\ldots, \\
     & \mathit{ite}(\pi.c=h-1,a,\pi.a_{h-1}),\mathit{ite}(\pi.c=h-1,o,\pi.o_h),a,o ) 
\end{split}
\end{equation*}

To encode the term $\pi = \omega$, we must take into account the lengths and make sure to ignore the unused elements in the tuple. Thus we encode $\pi = \omega$, assuming both have the same upper bound $h$, as:

\begin{equation*}
\begin{split}
\pi.c = \omega.c \wedge \bigwedge _{0 \leq i < \pi.c} (\pi.a_{i}=\omega.a_i \wedge \pi.o_{i+1}=\omega.o_{i+1})
\end{split}
\end{equation*}

A definition of a symbolic relation $B(\omega,s) \equiv \phi(\omega,s) \ | \ \omega \in \mathcal{H}_k, s \in S$, can be expressed as:
\[
B = \lambda \omega_k, s.\ \omega_k.c = k \wedge \phi(\omega_k,s)
\]
where $\omega_k$ is a fresh history variable with upper bound $h = k$.
This encoding is sufficient to implement the algorithms presented here, since all of the defined symbolic relations over histories have fixed upper bounds on the length. 

\section{Proofs}
\label{app:proofs}

\subsection{Proof of Theorem~\ref{thm:certificate-sound}}
Let $(s_i)_{i\ge 0}$ be an arbitrary induced run of $\Strat$ in $\game$.
By definition, there exists an associated memory sequence $(m_i)_{i\ge 0}$ such
that $m_0=\MemInit$ and, for every $i\ge 0$, we have:
\[
  \Model(s_i,\Policy(m_i),s_{i+1},\ObsFunc(s_{i+1}))
  \quad\text{and}\quad
  m_{i+1}=\MemUpdate(m_i,\ObsFunc(s_{i+1})).
\]
To show that $\Strat$ is winning, it suffices to prove that this arbitrary
induced run is safe, i.e., that $\Safe(s_i)$ holds for every $i\ge 0$. Using induction on $i$,  we prove
a stronger claim, namely: 
\[
\forall i \geq 0. \ \ \WinningMem(m_i) \wedge \SafeReach(m_i,s_i).
\]
By the \textsc{Safe} condition of the certificate, this  implies
$\Safe(s_i)$ at every position, meaning $\Strat$ is
winning.

  \noindent
{\bf Base case:} Since $s_0$ is initial, $\Init(s_0)$ holds. By the
\textsc{Init} condition of the certificate, $\WinningMem(\MemInit)$ holds and
$\Init(s_0)$ implies $\SafeReach(\MemInit,s_0)$. Since $m_0=\MemInit$, we have
$ 
  \WinningMem(m_0) \wedge \SafeReach(m_0,s_0)$. \\ 

  \noindent
{\bf Inductive step :}
By the inductive hypothesis, we have:
$
  \WinningMem(m_i) \wedge \SafeReach(m_i,s_i)$.
Let $o_{i+1}=\ObsFunc(s_{i+1})$. By definition of induced run, we also have:
\[
  \Model(s_i,\Policy(m_i),s_{i+1},o_{i+1})
  \quad\text{and}\quad
  m_{i+1}=\MemUpdate(m_i,o_{i+1}).
\]
Applying the \textsc{Step} condition of the certificate to
$m_i,s_i,s_{i+1},o_{i+1}$ yields
\[
  \WinningMem(\MemUpdate(m_i,o_{i+1}))
  \wedge
  \SafeReach(\MemUpdate(m_i,o_{i+1}),s_{i+1}).
\]
Using $m_{i+1}=\MemUpdate(m_i,o_{i+1})$, we conclude
$ 
  \WinningMem(m_{i+1}) \wedge \SafeReach(m_{i+1},s_{i+1})$.

\subsection{Proof of Theorem~\ref{thm:quotient-correctness}, Part 1: Soundness}

The proof relies on a  \emph{coupling invariant} relating the two games:
along any induced run of $\Strat$,  $\SafeReach(m_i,s_i)$ holds at every step.
Because $\SafeReach(m,s)\Rightarrow\Safe(s)$, this implies
 safety of the run. 

Let $(s_i)_{i\ge 0}$ be an arbitrary induced run of
$\Strat=\langle \Mem,\MemInit,\MemUpdate,\Policy\rangle$ in $\game$, with
associated memory sequence $(m_i)_{i\ge 0}$. By the definition of an induced
run, we have $\Init(s_0)$, $m_0=\MemInit$, and, 
\[
  \Model(s_i,\Policy(m_i),s_{i+1}, o_{i+1})
  \quad\text{and}\quad
  m_{i+1}=\MemUpdate(m_i,o_{i+1}),
\]
where $o_{i+1}=Obs(s_{i+1})$. We prove by induction on $i$ the following
invariant:
\begin{quote}
$I_i \quad \equiv\quad$
the prefix $m_0,o_1,m_1,\ldots,o_i,m_i$ is generated by $(\Policy,\MemUpdate)$,
and $\SafeReach(m_i,s_i)$ holds.
\end{quote}

\paragraph{Base case.} We prove $I_0$ in the base case.
The prefix $m_0$ is trivially generated. Since $\Init(s_0)$ holds and
$\MemInit$ satisfies
$\Init_{\SafeReach}(\MemInit)\equiv \forall s.\,\Init(s)\Rightarrow\SafeReach(\MemInit,s)$,
we obtain $\SafeReach(m_0,s_0)$ from $m_0=\MemInit$.

\paragraph{Inductive step.}
We show $I_{i+1}$ from inductive hypothesis $I_i$. To do so, we first show that the observation $o_{i+1}$ is enabled at
$m_i$. By definition,
\[
  \Enabled_{\SafeReach}\bigl(m_i,\Policy(m_i),o_{i+1}\bigr)
  \equiv
  \exists s,s'.\;
    \SafeReach(m_i,s)\,\wedge\,\Model(s,\Policy(m_i),s',o_{i+1}).
\]
Taking $s=s_i$ and $s'=s_{i+1}$, the first conjunct holds by the induction
hypothesis and the second by the induced run. Hence
$\Enabled_{\SafeReach}(m_i,\Policy(m_i),o_{i+1})$ holds. Together with
$m_{i+1}=\MemUpdate(m_i,o_{i+1})$, this shows that the extended prefix
$m_0,o_1,\ldots,o_{i+1},m_{i+1}$ is generated by $(\Policy,\MemUpdate)$.

Since $(\Policy,\MemUpdate)$ is winning from $\MemInit$, the {Update
validity} condition applies to the generated prefix ending in $m_i$: for every
observation $o$ with $\Enabled_{\SafeReach}(m_i,\Policy(m_i),o)$, we have
$\Model_{\SafeReach}(m_i,\Policy(m_i),o,\MemUpdate(m_i,o))$. Instantiating with
the enabled observation $o=o_{i+1}$ yields
$
  \Model_{\SafeReach}\bigl(m_i,\Policy(m_i),o_{i+1},m_{i+1}\bigr)$
where $m_{i+1}=\MemUpdate(m_i,o_{i+1})$. By the definition of the
induced transition relation,
\[
  \Model_{\SafeReach}(m,a,o,m')
  \equiv
  \forall s,s'.\;
    \SafeReach(m,s)\wedge\Model(s,a,s',o)\Rightarrow\SafeReach(m',s').
\]
Instantiating this with $s=s_i$, $s'=s_{i+1}$, $a=\Policy(m_i)$, $o=o_{i+1}$,
and $m'=m_{i+1}$, and using $\SafeReach(m_i,s_i)$ as the IH together
with $\Model(s_i,\Policy(m_i),s_{i+1},o_{i+1})$, we conclude
$\SafeReach(m_{i+1},s_{i+1})$, establishing $I_{i+1}$.
 \qed

\subsection{Proof of Theorem~\ref{thm:quotient-correctness}, Part 2: Relative completeness}

We first state and prove a  lemma that will also be useful for later proofs.

\begin{lemma}
\label{lem:winning-region}
Let $\Phi$ be a memory predicate and $(\Policy,\MemUpdate)$ a strategy for
$\game_{\SafeReach}$ such that:
\begin{description}[leftmargin=*]
  \item[(C1) Safety]   $\Phi(m)\Rightarrow\Safe_{\SafeReach}(m)$ for all $m$;
  \item[(C2) Inductivity]
  $\forall m,o,s,s'.\ \Phi(m)\wedge\SafeReach(m,s)\wedge\Model(s,\Policy(m),s',o)
   \ \Rightarrow\ \Phi(\MemUpdate(m,o))\wedge\SafeReach(\MemUpdate(m,o),s')$.
\end{description}
Then $(\Policy,\MemUpdate)$ is winning from \emph{every} memory $m_0$ with
$\Phi(m_0)$.
\end{lemma}

\begin{proof}
Consider any $m_0$ with $\Phi(m_0)$ and an arbitrary prefix
$m_0,o_1,m_1,\ldots,o_k,m_k$ generated by $(\Policy,\MemUpdate)$, so for every
$0\le i<k$, $\Enabled_{\SafeReach}(m_i,\Policy(m_i),o_{i+1})$ holds and
$m_{i+1}=\MemUpdate(m_i,o_{i+1})$. We first prove by induction on $i$ the
invariant
$ \Phi(m_i)$ holds.

\paragraph{Base case.} This is trivial since $\Phi(m_0)$ is assumed.

\paragraph{Inductive step.} We show $\Phi_{i+1}$ using the inductive hypothesis $\Phi_i$, for $i<k$. Since the
prefix is generated, $\Enabled_{\SafeReach}(m_i,\Policy(m_i),o_{i+1})$ holds; by
definition,
\[
  \Enabled_{\SafeReach}\bigl(m_i,\Policy(m_i),o_{i+1}\bigr)
  \equiv
  \exists s,s'.\;\SafeReach(m_i,s)\wedge\Model\bigl(s,\Policy(m_i),s',o_{i+1}\bigr),
\]
Let $s,s'$ be witnesses for the existential quantifier. Applying \textbf{(C2)} with $m=m_i$, $o=o_{i+1}$ and
these $s,s'$, and using $\Phi(m_i)$, yields
$\Phi(\MemUpdate(m_i,o_{i+1}))$, i.e., $\Phi(m_{i+1})$.

\paragraph{Safety.} By the invariant $\Phi(m_k)$ holds, so \textbf{(C1)} gives
$\Safe_{\SafeReach}(m_k)$.

\paragraph{Update validity.} Let $o$ satisfy
$\Enabled_{\SafeReach}(m_k,\Policy(m_k),o)$. We must show
$\Model_{\SafeReach}(m_k,\Policy(m_k),o,\MemUpdate(m_k,o))$, i.e.,
\[
  \forall s,s'.\;
    \SafeReach(m_k,s)\wedge\Model(s,\Policy(m_k),s',o)
    \Rightarrow\SafeReach(\MemUpdate(m_k,o),s').
\]
We consider arbitrary $s,s'$ with $\SafeReach(m_k,s)$ and $\Model(s,\Policy(m_k),s',o)$.
Since $\Phi(m_k)$ holds, \textbf{(C2)} (with $m=m_k$) yields
$\SafeReach(\MemUpdate(m_k,o),s')$. As $s,s'$ were arbitrary, the implication
holds.

Both conditions hold for every generated prefix from $m_0$; hence
$(\Policy,\MemUpdate)$ is winning from $m_0$.
\end{proof}

The relative completeness theorem now follows as a direct corollary of this lemma:

\begin{corollary}[Theorem~\ref{thm:quotient-correctness}, Part~2]
\label{cor:relative-completeness}
Suppose $\Strat=\langle \Mem,\MemInit,\MemUpdate,\Policy\rangle$ is a strategy for
$\game$, $(\SafeReach,\Phi)$ is an epistemic certificate for $\Strat$, and
$\Phi(m)\Rightarrow\Safe_{\SafeReach}(m)$. Then
$(\Policy,\MemUpdate)$ is winning from $\MemInit$ in $\game_{\SafeReach}$.
\end{corollary}

\begin{proof}
Condition \textbf{(C1)} is the assumed $\Phi\Rightarrow\Safe_{\SafeReach}$, and
\textbf{(C2)} is precisely the \textbf{(Step)} condition of the certificate. The
\textbf{(Init)} condition gives $\Phi(\MemInit)$, so
Lemma~\ref{lem:winning-region} applies at $m_0=\MemInit$.
\end{proof}

\subsection{Proof of Theorem~\ref{thm:findwinning-sound}}
\label{app:proof-findwinning-sound}

\paragraph{Strategy soundness.}
Suppose \textsc{FindWinning} returns a strategy
$ 
  \Strat = \langle \Mem,\MemInit,\MemUpdate,\Policy\rangle .
$
Based on Algorithm~\ref{alg:solve-po-game}, this can happen only if, in some
loop iteration, the algorithm constructs
the induced memory game $\game_{\SafeReach}$, and
$ 
  \SolveMemory(\game_{\SafeReach})
$
returns
$ 
  \mathit{Success}(\MemInit,\Policy,\MemUpdate).
$
By Proposition~\ref{prop:solve-memory-correct}, success of \textsc{SolveMemory}
implies that the memory-based strategy
$ 
  \langle \Mem,\MemInit,\MemUpdate,\Policy\rangle
$
is winning for the original partially observable game. Hence the strategy
returned by \textsc{FindWinning} is winning.

\paragraph{Refutational soundness.}
Suppose \textsc{FindWinning} returns \textsc{Failure}.  Based on 
Algorithm~\ref{alg:solve-po-game}, this can happen only if, in some iteration,
\textsc{InferMemory} returns $(\bot,\_)$. By Algorithm~\ref{alg:compute-safereach},
this happens  only when
$ 
  \textsc{ExploreBeliefs}(\Init,\Model,\Safe,h)
$ 
returns \textsc{Failure}. By Theorem~\ref{thm:explore-beliefs-correctness},
failure of \textsc{ExploreBeliefs} implies that the original game has no winning
strategy. Therefore, if \textsc{FindWinning} returns
\textsc{Failure}, then the original game has no winning strategy.

\subsection{Proof of Theorem~\ref{thm:conditional-success}}
\label{app:proof-conditional-success}

\rev{
Suppose that, at some iteration of
Algorithm~\ref{alg:solve-po-game},
\textsc{InferMemory} returns a memory domain $\Mem$
and an epistemic support relation $\SafeReach$
satisfying the theorem's assumptions.
In particular, there exists a strategy
$
  \Strat = \langle \Mem,\MemInit,\MemUpdate,\Policy\rangle
$
admitting an epistemic certificate $(\SafeReach,\Phi)$
such that $\Phi(m) \Rightarrow \phi_{\SafeReach}(m)$.

By the relative completeness part of
Theorem~\ref{thm:quotient-correctness},
$(\Policy,\MemUpdate)$ is a winning strategy for
the induced memory game $\game_{\SafeReach}$
from $\MemInit$. Furthermore, by the (Init)
condition of the certificate,
$I_{\SafeReach}(\MemInit)$ holds.

Let $\Phi^\star$ denote the winning region of
$\game_{\SafeReach}$, characterized by the greatest
fixpoint of the controllable predecessor operator.
Since $(\Policy,\MemUpdate)$ is winning from
$\MemInit$, we have $\Phi^\star(\MemInit)$.

By assumption, the fixpoint computation in
$\SolveMemory(\game_{\SafeReach})$ terminates.
Hence, the algorithm computes $\Phi^\star$.
Since
$
  I_{\SafeReach}(\MemInit) \land
  \Phi^\star(\MemInit)
$
holds, the algorithm does not return
\textsc{Failure}. Instead, it extracts a winning
strategy and returns
$
  \mathit{Success}(\MemInit',\Policy',\MemUpdate')
$
for some initial memory $\MemInit'$.
Thus, \textsc{FindWinning} returns
$
  \langle \Mem,\MemInit',\MemUpdate',\Policy'\rangle,
$
which is winning for the original game by
Theorem~\ref{thm:findwinning-sound}.
}

\subsection{Proof of Proposition~\ref{prop:solve-memory-correct}}

\begin{proof}
In both cases the algorithm has reached a fixpoint, so
$\WinningMem=\nu X.\,\mathsf{Pre}_{\SafeReach}(X)$.

\medskip\noindent\emph{(1) Soundness.}
We verify the two hypotheses of the winning-region lemma
(\Cref{lem:winning-region}) for $\WinningMem$ and $(\Policy,\MemUpdate)$. First,
the conjunct $\Safe_{\SafeReach}(m)$ in $\mathsf{Pre}_{\SafeReach}$ gives
$\WinningMem(m)\Rightarrow\Safe_{\SafeReach}(m)$, which is condition \textbf{(C1)}.
Second, $\textsc{ExtractStrategy}$ returns $(\Policy,\MemUpdate)$ satisfying
\[
\forall m,o.\
\WinningMem(m)\wedge\Enabled_{\SafeReach}(m,\Policy(m),o)
\Rightarrow
\WinningMem(\MemUpdate(m,o))\wedge\Model_{\SafeReach}(m,\Policy(m),o,\MemUpdate(m,o)),
\]
which is condition \textbf{(C2)}. By \Cref{lem:winning-region},
$(\Policy,\MemUpdate)$ is winning from every $m\in\WinningMem$. In the
$\mathit{Success}$ case, line~8 chose $\MemInit$ with
$\Init_{\SafeReach}(\MemInit)\wedge\WinningMem(\MemInit)$, so in particular
$(\Policy,\MemUpdate)$ is winning from $\MemInit$. Since
$\Init_{\SafeReach}(\MemInit)$ holds, \Cref{thm:quotient-correctness} (Soundness)
applies and yields that $\Strat=\langle\Mem,\MemInit,\MemUpdate,\Policy\rangle$ is
a winning strategy for $\game$.

\medskip\noindent\emph{(2) Relative completeness.}
For contradiction, suppose  $\game$ has a winning strategy
$\Strat'=\langle\Mem,\MemInit',\MemUpdate',\Policy'\rangle$ with an epistemic
certificate $(\SafeReach,\Phi^\star)$ such that $\Phi^\star\Rightarrow\Safe_{\SafeReach}$.

We first  show that $\Phi^\star$ is a \emph{post-fixpoint} of $\mathsf{Pre}_{\SafeReach}$, i.e.,
$\Phi^\star(m)\Rightarrow\mathsf{Pre}_{\SafeReach}(\Phi^\star)(m)$ for all $m$. To this end, 
consider $m$ satisfying $\Phi^\star(m)$. The conjunct $\Safe_{\SafeReach}(m)$ holds by
$\Phi^\star\Rightarrow\Safe_{\SafeReach}$. For the remaining conjunct, take the action
$a=\Policy'(m)$, and let $o$ be any observation with
$\Enabled_{\SafeReach}(m,\Policy'(m),o)$. By definition of $\Enabled_{\SafeReach}$
there exist $s,s'$ with $\SafeReach(m,s)\wedge\Model(s,\Policy'(m),s',o)$. Applying
the certificate's \textbf{(Step)} condition to $m$, $o$, and these $s,s'$ (using
$\Phi^\star(m)$) yields
\[
  \Phi^\star(\MemUpdate'(m,o))\ \wedge\ \SafeReach(\MemUpdate'(m,o),s').
\]
Moreover, \textbf{(Step)} holds for \emph{all} $s,s'$ compatible with $m$ and $o$,
so $\MemUpdate'(m,o)$ satisfies:
\[ \Model_{\SafeReach}(m,\Policy'(m),o,\MemUpdate'(m,o))
\]
which by definition is
$\forall s,s'.\,\SafeReach(m,s)\wedge\Model(s,\Policy'(m),s',o)\Rightarrow
\SafeReach(\MemUpdate'(m,o),s')$. Thus $m'=\MemUpdate'(m,o)$ witnesses
$\Phi^\star(m')\wedge\Model_{\SafeReach}(m,\Policy'(m),o,m')$ for every enabled $o$. Hence:
\[
  \exists a.\ \forall o.\
  \Enabled_{\SafeReach}(m,a,o)\Rightarrow
  \exists m'.\ \Phi^\star(m')\wedge\Model_{\SafeReach}(m,a,o,m').
\]
Along with $\Safe_{\SafeReach}(m)$ this is exactly
$\mathsf{Pre}_{\SafeReach}(\Phi^\star)(m)$, establishing that $\Phi^\star$ is a post-fixpoint.

Since $\WinningMem=\nu X.\,\mathsf{Pre}_{\SafeReach}(X)$ is the \emph{greatest} fixpoint, we have $\Phi^\star\Rightarrow\WinningMem$. Finally, the
certificate's \textbf{(Init)} condition provides $\MemInit'$ with $\Phi^\star(\MemInit')$
and $\forall s.\,\Init(s)\Rightarrow\SafeReach(\MemInit',s)$; the latter is exactly
$\Init_{\SafeReach}(\MemInit')$. Hence
$\Init_{\SafeReach}(\MemInit')\wedge\Phi^\star(\MemInit')$ holds, and since
$\Phi^\star\Rightarrow\WinningMem$, also
$\Init_{\SafeReach}(\MemInit')\wedge\WinningMem(\MemInit')$. This contradicts the
membership test on line~7.
\end{proof}

\subsection{Proof of Theorem~\ref{thm:explore-beliefs-correctness}}

\paragraph{Safety.}
Let $\widehat B_0,\ldots,\widehat B_h$ denote the beliefs computed by the
forward recurrence before the final filtering by $\mathsf{Win}$. We first show
that these beliefs contain only safe states:
\[
  \forall k\le h.\ \widehat B_k(\ohist,s)\Rightarrow\Safe(s).
\]
The proof is by induction on $k$. For $k=0$, we have
$ 
  \widehat B_0(\epsilon,s)\equiv \Init(s)\wedge\Safe(s),
$
so the claim holds. For the inductive step, consider a history
$\ohist\cdot a\cdot o$ of length $k$. By the belief recurrence,
\[
  \widehat B_k(\ohist\cdot a\cdot o,s')
  \equiv
  \Safe(s')\wedge
  \exists s.\,\widehat B_{k-1}(\ohist,s)\wedge\Model(s,a,s',o).
\]
Hence any state $s'$ satisfying
$\widehat B_k(\ohist\cdot a\cdot o,s')$ also satisfies $\Safe(s')$. Thus every
belief computed by the forward recurrence contains only safe states.

The beliefs returned by \textsc{ExploreBeliefs} are obtained from these beliefs
by conjoining the corresponding bounded-winning predicate:
\[
  B_k(\ohist,s) \equiv
  \widehat B_k(\ohist,s)\wedge\mathsf{Win}_k(\ohist).
\]
Therefore we have $B_k(\ohist,s)\Rightarrow\Safe(s)$.

\paragraph{Bounded refutation.}
We prove the contrapositive. Suppose the original game has a winning strategy
$\Strat=\langle\Mem,\MemInit,\MemUpdate,\Policy\rangle$. We show that
$\mathsf{Win}_0(\epsilon)$ must hold.

For a history $\ohist$ of length $k$, let $m_\ohist$ be the memory value obtained
by running $\Strat$ on $\ohist$ from the initial memory. We prove the following
claim by backward induction on $k$:

\[
  \text{If }\ohist\text{ is consistent with }\Strat
  \text{ and }\widehat B_k(\ohist,s)\text{ holds for some }s,
  \text{ then }\mathsf{Win}_k(\ohist)\text{ holds.}
\]

\vspace{0.1in}
At depth $h$, this holds because
$\mathsf{Win}_h(\ohist)\equiv\top$. Now suppose $k<h$, and assume the claim holds
for depth $k+1$. Let $\ohist$ be a length-$k$ history consistent with $\Strat$,
and let
$ 
  a = \Policy(m_\ohist)$
be the action chosen by the strategy after this history. We show that this
action witnesses the existential quantifier in the definition of
$\mathsf{Win}_k(\ohist)$.

Consider arbitrary $s,s',o$ such that
$ 
  \widehat B_k(\ohist,s)\wedge\Model(s,a,s',o)$.
By the belief recurrence, the successor state $s'$ belongs to the next belief $
  \widehat B_{k+1}(\ohist\cdot a\cdot o,s')
$
if $\Safe(s')$ holds. Since $\Strat$ is winning for the infinite-horizon
game, every transition that can arise from a state compatible with the current
history and under action $a$ must be safe; otherwise there would be an induced
run of $\Strat$ that violates safety. Hence $\Safe(s')$ holds. Moreover, the
extended history $\ohist\cdot a\cdot o$ is again consistent with $\Strat$, and
$\widehat B_{k+1}(\ohist\cdot a\cdot o,s')$ holds. By the induction hypothesis,
$
  \mathsf{Win}_{k+1}(\ohist\cdot a\cdot o)$
holds.

Hence,
if \textsc{ExploreBeliefs} returns \textsc{Failure}, then
$\mathsf{Win}_0(\epsilon)$ is false, and the original game has no winning strategy.

\section{Running Time Breakdown}
\label{app:running-breakdown}

\begin{figure}[t]
  \centering
\includegraphics[width=\linewidth]{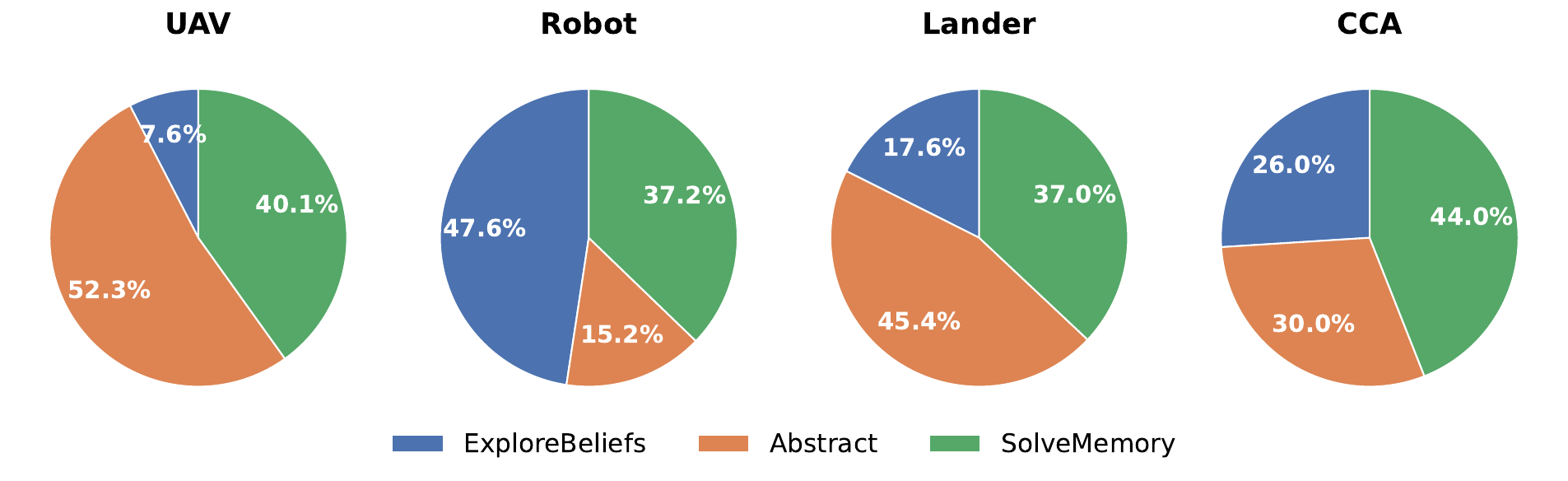}
\vspace{-0.1in}
  \caption{Time breakdown for \tool{} across key procedures in Algorithm~\ref{alg:solve-po-game}. The data represents the average distribution over  solved benchmarks.}
  \vspace{-0.2in}
  \label{fig:running-breakdown}
\end{figure}

This section provides a detailed breakdown of the running time of key
procedures in Algorithm~\ref{alg:solve-po-game}.

Figure~\ref{fig:running-breakdown}
shows the relative time spent in each phase across the four benchmark
families. Across all benchmarks, \SolveMemory{} occupies a substantial share of the total computation time. In particular, the percentage of \SolveMemory{} time ranges from $37.0\%$ (Lander) to $44.0\%$ (CCA), reflecting the
persistent cost of solving the induced perfect-information game.

The cost of \ExploreBeliefs{} varies substantially by domain, correlating
with model complexity. Among all benchmarks, \ExploreBeliefs{} takes the largest share in   the \textbf{Robot} family, taking $47.6\%$ of the running time. In contrast, it has the smallest share in \textbf{UAV} (7.6\%).   This difference can be attributed to   the relative complexity of the
environment models. In particular, the transition relations in the Robot benchmarks are
substantially larger than in the other three domains, and larger models
directly inflate the cost of belief exploration.

\Abstract{} is the dominant phase for \textbf{UAV} ($52.3\%$) and remains a
significant component for \textbf{Lander} ($45.4\%$), indicating that mining
concise epistemic supports from history-based beliefs incurs a non-trivial
cost. This is considerably more than the other benchmark families because the memory
templates for \textbf{UAV} and \textbf{Lander} contain a large number of
memory variables, which substantially increases the cost of finding
epistemic supports. 

\fi

\end{document}